%% file: main.tex
\documentclass[sigconf]{acmart}
\PassOptionsToPackage{table,xcdraw}{xcolor}  
\usepackage{xcolor}  

\usepackage{multirow}
\usepackage{tabularx} 
\usepackage{amsmath}

\usepackage{amssymb}
\usepackage{float}
\usepackage{algorithm}
\usepackage{colortbl}
\usepackage{algpseudocode}  
\usepackage{booktabs}
\usepackage{appendix}
\usepackage{graphicx}
\usepackage{subcaption}
\usepackage{array}
\usepackage[absolute,overlay]{textpos}

\AtBeginDocument{%
  \providecommand\BibTeX{{%
    \normalfont B\kern-0.5em{\scshape i\kern-0.25em b}\kern-0.8em\TeX}}}

\copyrightyear{2025}
\acmYear{2025}
\setcopyright{rightsretained}
\acmConference[CIKM '25] {Proceedings of the 34th ACM International Conference on Information and Knowledge Management}{ November 10--14, 2025}{Seoul, Republic of Korea.}
\acmBooktitle{Proceedings of the 34th ACM International Conference on Information and Knowledge Management (CIKM '25), November 10--14, 2025, Seoul, Republic of Korea}
\acmISBN{979-8-4007-2040-6/2025/11}
\acmDOI{10.1145/3746252.3761099}

\begin{document}

\title{Leveraging Multi-facet Paths for Heterogeneous Graph Representation Learning}


\author{Jongwoo Kim}
\affiliation{%
  \institution{Department of Industrial \& Systems Engineering, KAIST}
  \city{Daejeon}
  \country{South Korea}
}
\email{gsds4885@kaist.ac.kr}

\author{Seongyeub Chu}
\affiliation{%
  \institution{Graduate School of Data Science, KAIST}
  \city{Daejeon}
  \country{South Korea}
}
\email{chseye7@kaist.ac.kr}

\author{Hyeongmin Park}
\affiliation{%
  \institution{Department of Industrial \& Systems Engineering, KAIST}
  \city{Daejeon}
  \country{South Korea}
}
\email{mike980406@kaist.ac.kr}

\author{Bryan Wong}
\affiliation{%
  \institution{Graduate School of Data Science, KAIST}
  \city{Daejeon}
  \country{South Korea}
}
\email{bryan.wong@kaist.ac.kr}

\author{Keejun Han}
\affiliation{%
  \institution{School of Computer Engineering, Hansung University}
  \city{Seoul}
  \country{South Korea}
}
\email{keejun.han@hansung.ac.kr}

\author{Mun Yong Yi}
\affiliation{%
  \institution{Department of Industrial \& Systems Engineering, KAIST}
  \city{Daejeon}
  \country{South Korea}
}
\email{munyi@kaist.ac.kr}
\authornote{Corresponding author.}

\begin{abstract}
Recent advancements in heterogeneous GNNs have enabled significant progress in embedding nodes and learning relationships across diverse tasks. However, traditional methods rely heavily on meta-paths grounded in node types, which often fail to encapsulate the full complexity of node interactions, leading to inconsistent performance and elevated computational demands. To address these challenges, we introduce MF2Vec, a novel framework that shifts focus from rigid node-type dependencies to dynamically exploring shared facets across nodes, regardless of type. MF2Vec constructs multi-faceted paths and forms homogeneous networks to learn node embeddings more effectively. Through extensive experiments, we demonstrate that MF2Vec achieves superior performance in node classification, link prediction, and node clustering tasks, surpassing existing baselines. Furthermore, it exhibits reduced performance variability due to meta-path dependencies and achieves faster training convergence. These results highlight its capability to analyze complex networks comprehensively. The implementation of MF2Vec is publicly available at \texttt{https://github.com/kimjongwoo-cell/MF2Vec}.

\end{abstract}

\ccsdesc[500]{Information systems~Data mining}
\ccsdesc[300]{Computing methodologies~Learning latent representations}
\ccsdesc[300]{Computing methodologies~Neural networks}

\keywords{Heterogeneous graph neural network, Graph-representative learning, Metapath, Multi-facet}

\maketitle

\input{1.Introduction_short} 
\input{3.Preliminary_short}

\input{4.Method_short}

\input{5.Experiment_short}

\input{6.Ablation}

\input{8.Casestudy}
\input{2.Relatedwork_short}

\input{7.Conclusion}

\bibliographystyle{ACM-Reference-Format}
\balance
\bibliography{main}

\end{document}

%% file: 1.Introduction_short.tex
\section{Introduction}

Graph Neural Networks (GNNs) have emerged as powerful tools for capturing complex structures and relationships within graph data, with applications spanning numerous domains \cite{kipf, hamilton, velickovic}. As a specialized branch of GNNs, Heterogeneous Graph Neural Networks (HGNNs) \cite{mecch, graphhinge, hmsg, magnn, han} aim to address the unique challenges of modeling graphs with diverse types of nodes and edges by introducing tailored mechanisms for efficient learning. Among these, relation-based HGNNs enhance learning by aggregating messages along edges categorized by relation types, yet they often encounter limitations due to the deep network structures required to process multi-hop relations, leading to over-smoothing and diminished performance \cite{mecch}.

\begin{figure}
    \centering
    \includegraphics[width=0.9\linewidth]{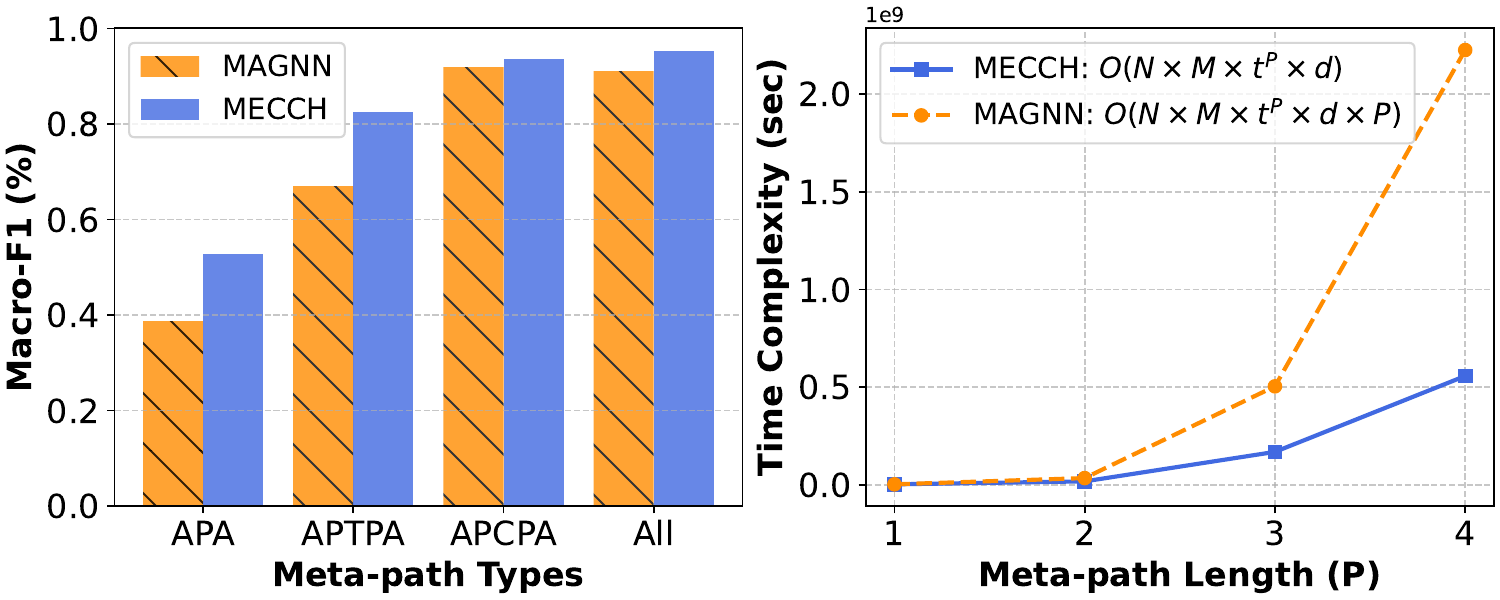}
    \Description{Plot showing Macro-F1 and time complexity comparison between MECCH and MAGNN on DBLP dataset.}
    \caption{%
        Performance (Macro-F1) and time complexity of MECCH and MAGNN on DBLP. 
        Left: Macro-F1 varies by meta-path, showing sensitivity to meta-path choice. 
        Right: Time complexity rises with meta-path length due to increased intermediate nodes. 
        $N$ (nodes), $M$ (meta-paths), $t$ (avg. neighbors), $P$ (meta-path length), $d$ (embedding dimension).
    }
    \label{fig:1_1}
\end{figure}

To mitigate these challenges, meta-path-based HGNNs leverage meta-paths to construct homogeneous subgraphs, reducing the need for multi-hop message passing. For instance, the ``MAM'' meta-path on the IMDB dataset links Movie and Actor nodes, capturing collaborative relationships between movies via shared actors \cite{mecch, magnn, wang2021self}. Despite their efficiency, such meta-path-based approaches are constrained by rigid node-type combinations, which give rise to several critical limitations.

First, meta-paths can only reflect specific relationships between certain types of nodes, which limits their ability to capture a wide range of relationships in a detailed manner. For example, the ``MAM'' meta-path captures collaborations between movies that share the same actor but fails to represent more detailed relationships beyond it such as sharing genres or countries. Second, as depicted in Figure~\ref{fig:1_1} (left), the performance of meta-path-based models fluctuates depending on the selected meta-path, resulting in instability across datasets and tasks \cite{mhnf, hu2020heterogeneous}. Third, as shown in Figure~\ref{fig:1_1} (right), the exponential increase in meta-path combinations and lengths with diverse intermediate nodes amplifies computational costs and exploration complexity. Even with advanced learnable meta-path approaches like GTN \cite{GTN} and MHNF \cite{mhnf}, these issues persist due to their reliance on fixed node-type configurations.

This paper introduces a novel Multi-Facet Path approach to address these limitations. The concept of a facet, representing distinct attributes or properties of nodes (e.g., filming location or production year for a movie), is expanded from prior works on homogeneous graphs \cite{liu, park2020unsupervised, r-GAT} to encompass shared characteristics across diverse node types in heterogeneous graphs. By shifting from node-type-based paths to facet-based constructions, the proposed method offers a flexible and granular modeling framework that overcomes the rigidity and inefficiency of traditional meta-path approaches.

Through dynamic facet embeddings, which are learned and updated in an end-to-end manner, the Multi-Facet Path approach enables the representation of intricate relationships across nodes, enhancing both efficiency and model stability. First, the use of facets allows for the capture of a broader range of relationships, enriching downstream task representations. Second, this method minimizes performance variability by dynamically constructing paths based on facet information rather than fixed meta-paths. Third, the elimination of manual meta-path definitions and the reduction of computational overhead further bolster scalability.

\begin{figure}
    \centering
    \includegraphics[width=\linewidth, trim=0 0 0 250, clip]{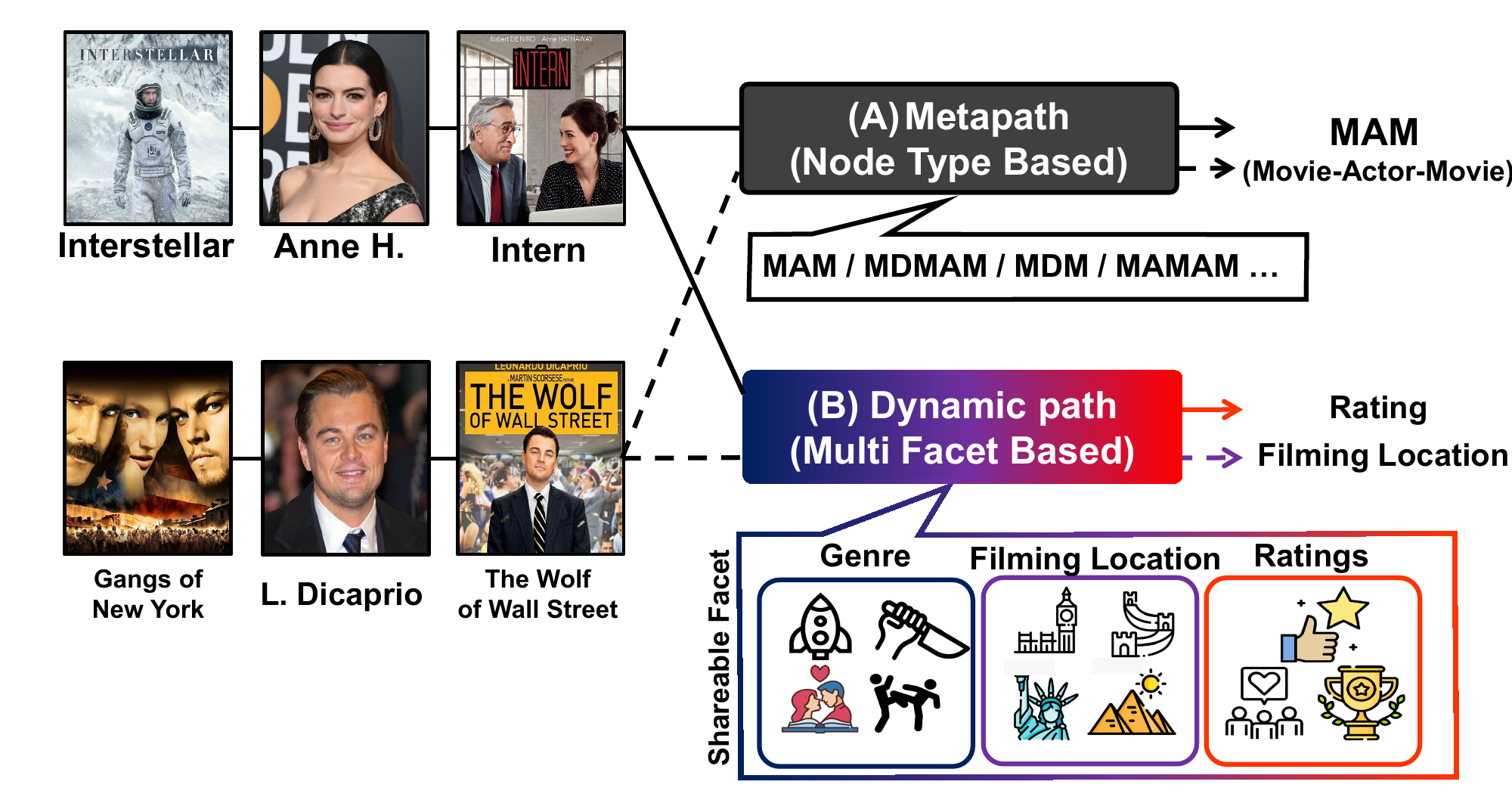}
    \Description{Schema comparison on IMDB data: (A) predefined node-type schemas vs. (B) dynamic multi-facet path assignment.}
    \caption{%
        Comparison of schema assignment for IMDB data. 
        (A) Existing methods use predefined schemas based on node types. 
        (B) The proposed approach dynamically assigns paths using multi-facet criteria, 
        selecting the most descriptive facet, such as filming location, genre, or rating.
    }
    \label{fig:1}
\end{figure}

Figure~\ref{fig:1} shows that when following the same meta-path, nodes can be grouped based on different facets. For example, the paths ``New York--L.\ Dicaprio--The Wolf of Wall Street'' and ``Interstellar--Anne H.--Intern'' may simply be grouped as \textit{``Movie--Actor--Movie''} under meta-path-based methods. However, with a multi-facet approach, the path ``New York--L.\ Dicaprio--The Wolf of Wall Street'' can be considered from the perspective of filming location, while the path ``Interstellar--Anne H.--Intern'' may reflect similarities in IMDb ratings. This approach allows the model to comprehensively capture latent and rich information within the graph, effectively overcoming the limitations of traditional meta-path-based methods.

In MF2Vec, paths are extracted from the adjacency matrix of a heterogeneous graph through random walks without distinguishing them by predefined meta-path types. The node embeddings of the extracted paths are projected into a shared facet space of $K$ dimensions. Gumbel Softmax and an attention mechanism are employed to select the most relevant facets for intermediate nodes, ensuring a discrete yet differentiable selection process. These facets characterize the relationships between endpoint nodes and compute path weights that reflect their importance for node connections. Finally, a homogeneous network is constructed by integrating same-typed nodes and fed into a GNN, which aggregates multi-facet vectors along the paths to generate latent node representations.

Key contributions of this work are summarized as follows:

\begin{itemize}
    \item \textbf{Dynamic Facet Embedding:} A novel framework for constructing flexible paths independent of node types.
    \item \textbf{Stable Performance:} Significant improvement in stability across datasets and tasks by mitigating dependence on fixed meta-paths.
    \item \textbf{Efficiency and Performance Enhancement:} Reduced computational costs and improved performance for node classification, node clustering, and link prediction tasks.
\end{itemize}

%% file: 3.Preliminary_short.tex
\section{Preliminary}
\noindent\textbf{Definition 1 (Heterogeneous Graph).} A heterogeneous graph \( G = (V, \mathcal{E}) \) is defined by a node type mapping function \( \phi : V \rightarrow \mathcal{A} \) and an edge type mapping function \( \psi : \mathcal{E} \rightarrow \mathcal{R} \), where \( \mathcal{A} \) and \( \mathcal{R} \) are sets of node and edge types, respectively, with \( |\mathcal{A}| + |\mathcal{R}| > 2 \).

\noindent\textbf{Definition 2 (Metapath).} A metapath is a sequence \( A_1 \xrightarrow{R_1} A_2 \xrightarrow{R_2} \cdots \xrightarrow{R_k} A_{k+1} \) (or \( A_1A_2 \cdots A_{k+1} \)), representing a composite relation \( R = R_1 \circ R_2 \circ \cdots \circ R_k \) between node types \( A_1 \) and \( A_{k+1} \).

\noindent\textbf{Definition 3 (Metapath Schema).}
A metapath schema is a sequence of node types $A_1, A_2, \dots, A_{k+1}$ connected by edge types $R_1, R_2, \dots, R_k$. This represents the structural pattern of a metapath and illustrates how different node types are related through specific types of relationships defined by the nodes and edges.

\noindent\textbf{Definition 4 (Multi-facet Node Representation Learning).} For a graph \( G = (V, \mathcal{E}) \) with nodes \( V \), the goal is to learn a multi-facet node embedding \( E_{facet}(v_i) \in \mathbb{R}^{K \times d} \) for each node \( v_i \). Here, \( K \) is the number of facets, and each facet embedding \( E_n(v_i) \in \mathbb{R}^{1 \times d} \) aims to: 1) preserve network structure, 2) capture various facets of \( v_i \), and 3) represent facet interactions.

\noindent\textbf{Definition 5 (Multi-facet Path).} Unlike predefined metapaths, a multi-facet path \( P = v_1 \xrightarrow{R_1} v_2 \xrightarrow{R_2} \cdots \xrightarrow{R_{m}} v_{m+1} \) is extracted without manual specification. Intermediate nodes \( v_2, \ldots, v_m \) are projected onto multi-facet vectors \( F_2, \ldots, F_m \), forming a multi-facet path \( P = v_1 \rightarrow F_2 \rightarrow \ldots \rightarrow F_m \rightarrow v_{m+1} \). This path integrates information from intermediate nodes through multiple facets, updating node representations based on various latent information beyond node types.

%% file: 4.Method_short.tex
\begin{figure*}
    \centering
    \includegraphics[width=0.8\linewidth, trim=0 80 0 0, clip]{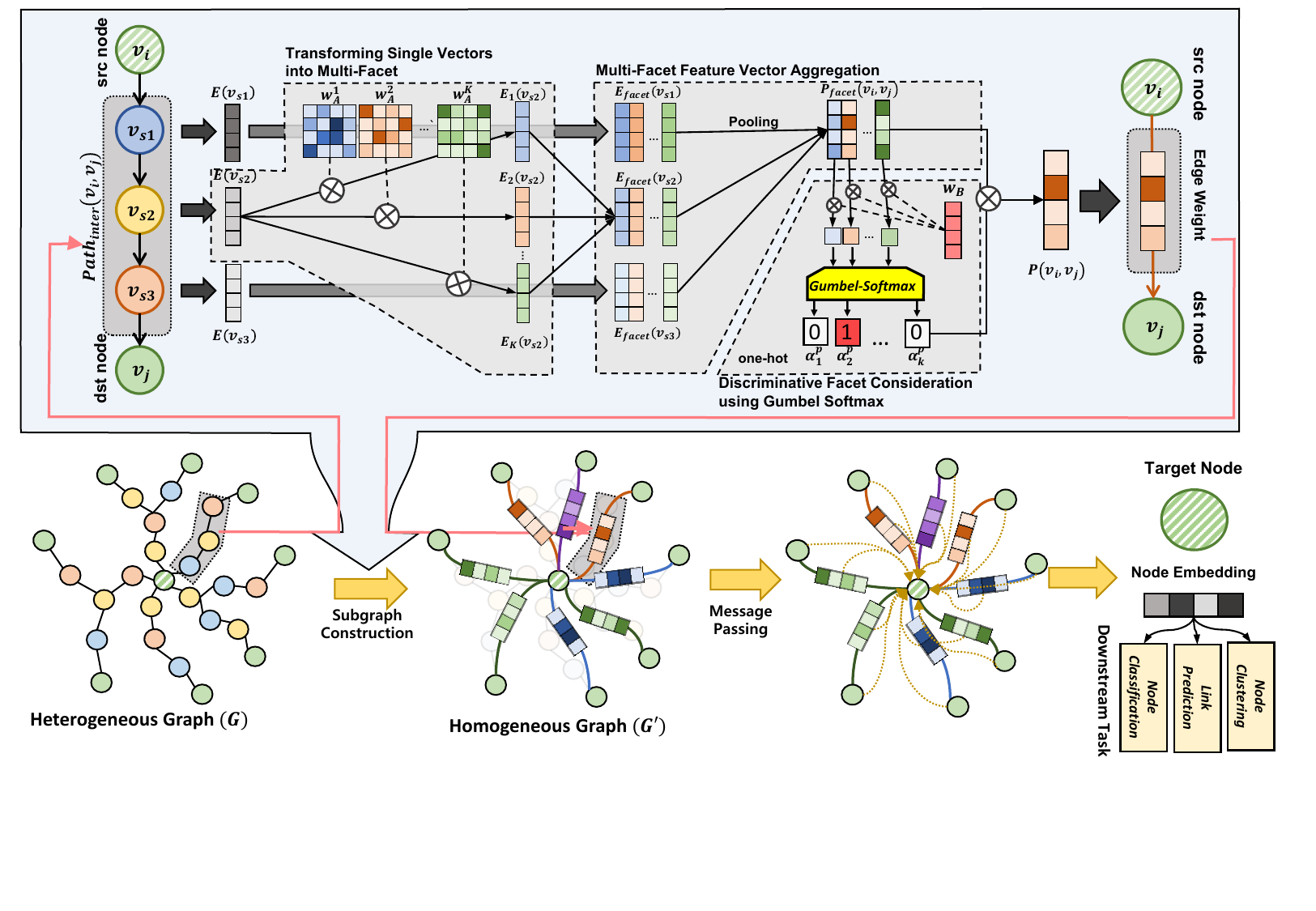}
    \Description{Overall architecture diagram of the MF2Vec model pipeline.}
    \caption{Overall Architecture}
    \label{method}
\end{figure*}

\begin{algorithm}[h]
\footnotesize
\caption{MF2Vec Forward Propagation}
\label{mf2vec_al}
\begin{algorithmic}
    \State \textbf{Input:} Heterogeneous graph $G$; initial embedding matrix $E_n(V') \in \mathbb{R}^{|V'| \times d}$
    \State \textbf{Output:} Final node embeddings $h^L(V')$
    \State Let $v_s$ denote intermediate nodes along the path $paths(v_i, v_j)$

    \State // \textbf{GNN Warm-up}
    \Repeat
        \State For each $v_i$ in $V'$:
            \State \hspace{1em} Compute $L_{\text{warm-up}}(v_i)$ using run-warm-up
        \State Let $L_{\text{warm-up}}(V') = \sum_{i=1}^{|V|} L_{\text{warm-up}}(v_i)$
        \State Update $E(V')$ by minimizing $L_{\text{warm-up}}$
    \Until{convergence}

    \State Construct subgraph $G' = (V', \mathcal{E}')$ via random walk among selected node types

    \State // \textbf{MF2Vec Propagation}
    \State For each $v_i$ in $V'$:
        \State \hspace{1em} Let $E_{\text{facet}}(v_i) = \{ W_1 E(v_i) \}_{n=1}^k$
        \State \hspace{1em} For each $(v_i, v_j)$:
        \State \hspace{2em} Let $P_{\text{facet}}(v_i, v_j) = \{ \text{Agg}(E_{\text{facet}}(v_s)) \}_{n=1}^k$
        \State \hspace{2em} Let $P(v_i, v_j) = \sum_{n=1}^k \sigma(W_2 P_n(v_i, v_j)) \cdot P_n(v_i, v_j)$

    \State For $l = 1$ to $L$:
        \State \hspace{1em} For each $v_i$ in $V'$:
        \State \hspace{2em} $h^{l+1}(v_i) = \sigma \left( \mathrm{BN} \left( \sum_{v_j \in N(i)} P(v_i, v_j) \cdot h^l(v_j) \right) \right)$

    \State \Return $h^L(V')$
\end{algorithmic}
\end{algorithm}

\section{Methodology}

In this section, we describe our approach, as shown in Figure~\ref{method}. We begin by extracting paths between homogeneous end nodes using random walks, similar to traditional metapath methods \cite{metapath2vec, hin2vec}. We then project the features of intermediate nodes into multiple facets and aggregate these facets to create multi-facet embedding representations. These representations are used as edge features between path-guided neighbors and the target node. The model pseudocode is provided in Algorithm~\ref{mf2vec_al}, and the notation is summarized in Table~\ref{notations}.

\begin{table}[!ht]
\caption{Notations}
\label{notations}
\centering
\renewcommand{\arraystretch}{1.1}
\resizebox{\columnwidth}{!}{%
\begin{tabular}{@{}ll@{}}
\toprule
Notation        & Explanation                                                                \\ \midrule
$\mathcal{R}^d$           & $d$-dimensional Euclidean space \\
$G=(V,\mathcal{E})$       & A heterogeneous graph with a node set $V$ and an edge set $\mathcal{E}$    \\
$G'= (V', \mathcal{E'})$  & A subgraph with a certain node type set $V'$ and a certain node type edge set $\mathcal{E'}$ \\
$v_i$           & A node $v_i \in V'$  \\
$E(v_i)$        & The embedding of node $v_i$     \\
$E_n(v_i)$      & $n$-th multi-facet embedding of node $v_i$\\
$E_{\text{facet}}(v_i)$ & Multi-facet embedding of node $v_i$ \\
$h^L(v_i)$      & Final embedding of node $v_i$ after $L$-th layer \\
$path(v_i, v_j)$    & The set of whole nodes in the path between node $v_i$ and $v_j$ \\
$path_{inter}(v_i, v_j)$ & Intermediate nodes in the path between node $v_i$ and $v_j$ \\
$P(v_i, v_j)$   & The embedding of the path between node $v_i$ and $v_j$     \\
$P_n(v_i, v_j)$ & $n$-th multi-facet embedding of the path between node $v_i$ and $v_j$     \\  
$P_{\text{facet}}(v_i, v_j)$ & Multi-facet embedding of the path between node $v_i$ and $v_j$ \\
$\alpha_n^p$    & Weight of $n$-th facet in the path $p$    \\ \bottomrule
\end{tabular}%
}
\end{table}

\subsection{Extracting Paths Using Random Walk}

Given a node type \( A \), our method constructs a homogeneous subgraph that identifies direct connections between nodes of type \( A \), the same as metapath-based models. Instead of relying on metapath schemas such as APA or APCPA, the method generates a path starting from a node of type \( A \) and applies a random walk, traversing randomly selected adjacent nodes until encountering a type A node. This approach captures relationships within a specified path length limit, effectively mapping connections within the nodes of a type without being constrained by predefined metapath patterns. Details of the \textit{Random Walk} algorithm are provided in Algorithm~\ref{algo1}.

\begin{algorithm}[!h]
\footnotesize
\caption{Random Walk}
\label{algo1}
\begin{algorithmic}
    \State \textbf{Input:} Heterogeneous graph $G$; node types $V'$; start node $v_i \in V'$; path length $l$
    \State \textbf{Output:} Random walk path $path(v_i, v_j)$ or None
    \State Initialize $path$ with $v_i$
    \State For $l$ steps:
    \State \hspace{1em} Let $v_{curr}$ be the last node in $path$
    \State \hspace{1em} Let $N_{curr}$ be the neighbors of $v_{curr}$ in $G$
    \State \hspace{1em} If $N_{curr}$ is empty: \Return None
    \State \hspace{1em} Randomly select $v_{next}$ from $N_{curr}$
    \State \hspace{1em} Add $v_{next}$ to $path$
    \State Let $v_j$ be the last node in $path$
    \State If $v_j \in V'$: \Return $path(v_i, v_j)$
    \State Else: \Return None
\end{algorithmic}
\end{algorithm}

\subsection{Transforming Single Embeddings into Multi-Facet}

Our model captures inherent information from all node types before generating a facet embedding. Node information is exchanged through a GNN warm-up phase, which facilitates the aggregation of neighborhood information and enriches the initial node embeddings with structural and semantic context. The node embedding, \(E(v_i)\), is defined during this warm-up phase and serves as the foundation for training the main model. The resulting node embedding \(E(v_i)\) is then refined into a multi-dimensional embedding \(E_{\text{facet}}(v_i)\) to capture diverse information, as detailed in Equation~\ref{embedding}.

\begin{equation}
    \mathbf{\textit{E}}_{\text{facet}}(v_i) = [\, \mathbf{\textit{E}}_1(v_i) \; || \; \ldots \; || \; \mathbf{\textit{E}}_K(v_i) \,], \quad
    \mathbf{\textit{E}}_n(v_i) = \mathbf{\textit{W}}_A^n \mathbf{\textit{E}}(v_i)
    \label{embedding}
\end{equation}

Here, $\mathbf{\textit{E}}_n(v_i)$ denotes the embedding of node $v_i$ for the $n$-th facet, where $n = 1, 2, \ldots, K$, $\mathbf{\textit{W}}_A^n$ is a learnable weight matrix of size $d \times d$ for each facet, and $K$ is the total number of facets. The embeddings for all facets are then concatenated to form a unified multi-facet embedding, $\mathbf{\textit{E}}_{\text{facet}}(v_i)$, providing a comprehensive representation of the node.

\subsection{Multi-facet Feature Embedding Aggregation}

To construct a multi-facet feature embedding for paths consisting of multiple nodes, we employ an aggregation function that synthesizes information from the nodes along the path. The mathematical formulation for this aggregation is given by Equation~\ref{path_embedding}.

\begin{align}
\mathbf{\textit{P}}_{\text{facet}}(v_i,v_j) = f(\mathbf{\textit{E}}_{\text{facet}}(v_s) \mid v_s \in \text{path}_{inter}(v_i,v_j))
\label{path_embedding}
\end{align}

In the given equation, $\mathbf{\textit{P}}_{\text{facet}}(v_i,v_j)$ represents the aggregated multi-facet feature embedding for a connection between nodes $v_i$ and $v_j$. The set $\text{path}_{inter}(v_i,v_j)$ consists of intermediate nodes between $v_i$ and $v_j$, and $v_s$ is an intermediate node within the path. The function $f$ signifies an aggregation operation using mean pooling.

\subsection{Discriminative Facet Consideration using Gumbel Softmax}

We utilize the Gumbel Softmax \cite{jang2016} function to introduce randomness and uncertainty into facet selection, which helps the model explore diverse facet influences in complex relationships. This approach enables us to capture nuanced facet interactions and dependencies, making our model more adaptable and robust for multi-facet data. Even if the random walk algorithm extracts the paths that are little related to the end nodes, we mitigate their impact by assigning higher weights to the most crucial components using Gumbel Softmax.

\begin{align}
\mathbf{\textit{P}}(v_i,v_j) = \sum_{n=1}^{K} \alpha^p_n \cdot \mathbf{\textit{P}}_{n}(v_i,v_j), \quad \alpha^p_n = \sigma(\mathbf{\textit{W}}_B\mathbf{\textit{P}}_{n}(v_i,v_j))
\end{align}

In the given equation, $\sigma$ represents Gumbel Softmax. $\textit{P}(v_i,v_j)$ is the final embedding for $path(v_i,v_j)$ that combines these probabilities, and $\textit{W}_B$ is a matrix with dimensions $\textit{1} \times \textit{d}$.

\subsection{Subgraph Construction}

\begin{figure}[!htp]
    \centering
    \includegraphics[width=0.75\linewidth, trim=0 0 300 0, clip]{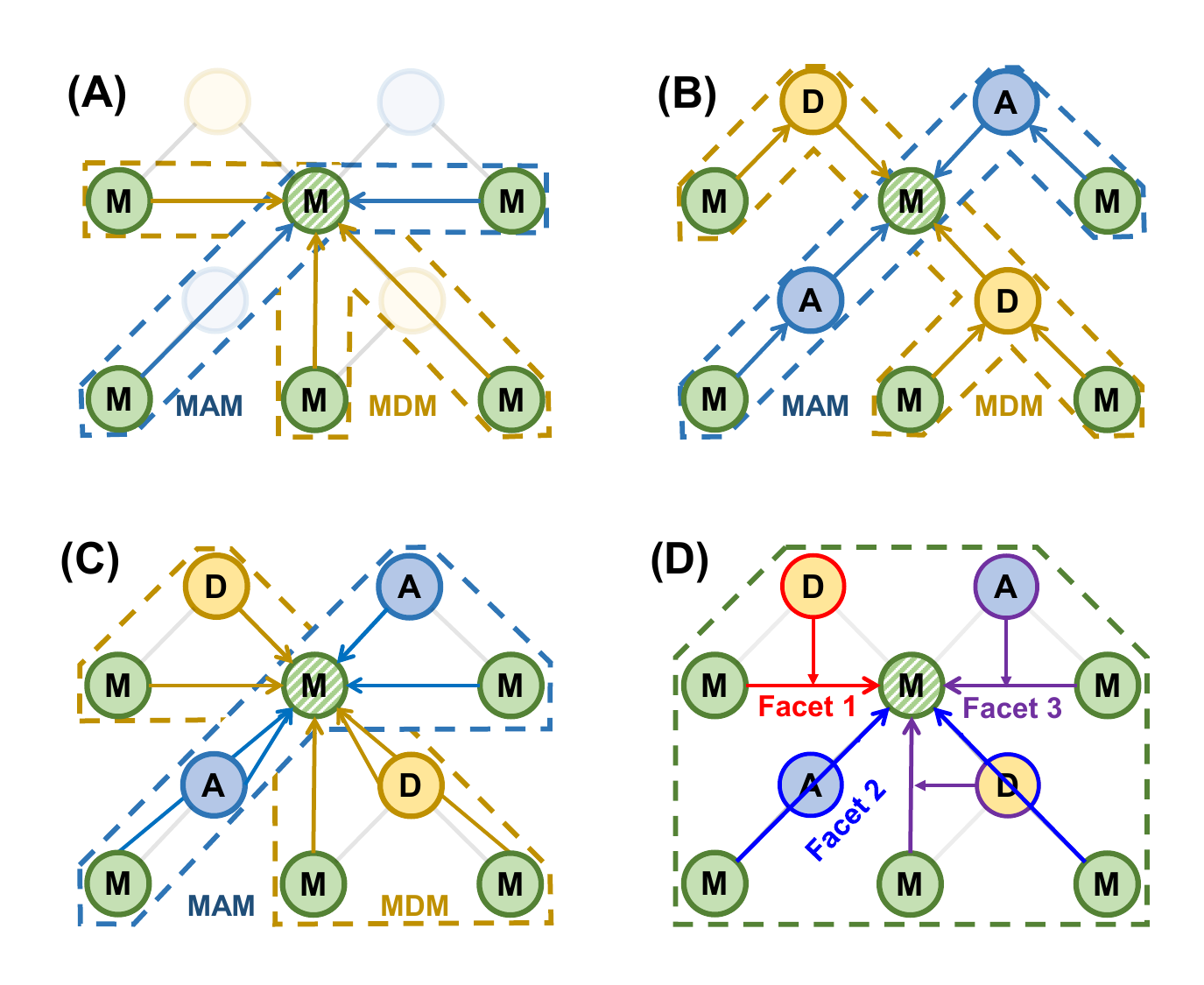}
    \Description{Comparison of HGNN aggregation strategies: HAN, MAGNN, MHNF, MECCH, and MF2Vec for a green target node.}
    \caption{HGNN aggregation for the dashed green target node, where M, D, and A denote node types. 
    (a) HAN: connects metapath-based neighbors, ignoring intermediate nodes. 
    (b) MAGNN \& MHNF: connect neighbors via metapaths, including intermediates. 
    (c) MECCH: directly connects all nodes along a metapath for context. 
    (d) MF2Vec: links multi-facet neighbors using facet-based edge features, independent of metapaths.}
    \label{fig:enter-label1}
\end{figure}

Path-based HGNNs transform the input graph into a new homogeneous subgraph $G^P=(V^P, \mathcal{E}^P)$ for each node and path, where $G^P$ varies by model. For example, in HAN, $G^P$ represents the metapath-$P$-guided neighborhood, while MAGNN defines it as instances of metapath-$P$ incident to the node. MECCH constructs subgraphs using most metapaths as illustrated in Figure~\ref{fig:enter-label1}.

Our approach, however, avoids the need for multiple metapaths by leveraging multi-facet information from a single path between nodes. This results in a subgraph $G'=(V', \mathcal{E}')$ that provides richer information than traditional methods. The MF2Vec model efficiently learns complex node relationships by constructing subgraphs incorporating multi-dimensional edge features which are multi-facet embedding for a more comprehensive node representation. Details are provided in Algorithm~\ref{algorithm_subgraph_generation}.

\begin{algorithm}[h]
\footnotesize
\caption{Sub-graph Generation via Random Walk Paths}
\label{algorithm_subgraph_generation}
\begin{algorithmic}
    \State \textbf{Input:} Heterogeneous graph $G$; node types $V'$; path length $l$; number of walks $n_{walks}$
    \State \textbf{Output:} Sub-graph $G' = (V', \mathcal{E}')$
    \State Initialize $paths \leftarrow []$, $pairs \leftarrow \emptyset$
    \State For each $v_i \in V'$ and for $n_{walks}$ times:
    \State \hspace{1em} $path \leftarrow$ RandomWalk$(G, V', v_i, l)$
    \State \hspace{1em} If $path \neq$ None:
    \State \hspace{2em} $pair \leftarrow (path[0], path[-1])$
    \State \hspace{2em} If $pair \notin pairs$: add $pair$ to $pairs$; add $path$ to $paths$
    \State Build $G'$ from $paths$ and return $G'$
\end{algorithmic}
\end{algorithm}

\subsection{Message Passing}

To extract node embeddings, we use GCN, considering node interactions and incorporating path facets as edge features. The graph node embedding is computed as follows:

\begin{align}
h^{(l+1)}(v_i) = \sigma\left(\text{BN}\left(\sum_{j\in \mathcal{N}(i)} \mathbf{\textit{P}}(v_i,v_j) \cdot h^{l}(v_j)\right)\right)
\end{align}

In this equation, \(\sigma\) represents the activation function, which is ELU, and BN represents the batch normalization. The \(P(v_i,v_j)\) denotes the path facet feature between node \(i\) and its neighbor \(j\), which is utilized as an edge feature.

\subsection{Training Losses}

We jointly train our model for node classification and link prediction. For node classification, we adopt the cross-entropy loss \(L_{nc} = -\sum_{v \in V_L} y_v \log(y'_v)\), where \(V_L\) is the set of labeled nodes, \(y_v\) is the true label, and \(y'_v\) is the predicted probability. For link prediction, we use a contrastive loss \(L_{lp} = -\sum_{(v, \mathrm{pos}) \in V^+} \log(\sigma(h_v^T h_{\mathrm{pos}})) - \sum_{(v, \mathrm{neg}) \in V^-} \log(\sigma(-h_v^T h_{\mathrm{neg}}))\), where \(V^+\) and \(V^-\) are the positive and negative node pairs, respectively, and \(\sigma(\cdot)\) denotes the sigmoid function. These two objectives enable the model to learn both node-level classification and structural link prediction simultaneously.

%% file: 5.Experiment_short.tex
\section{Experiment}

In this part, we assess the performance and practicality of our introduced method through tests on link prediction, node classification, and node clustering tasks across five diverse graph datasets. 



\noindent\textbf{Datasets.} We evaluate our model on six heterogeneous graph datasets: DBLP, ACM, IMDB, and Freebase for node classification and clustering, and Yelp and MovieLens for link prediction. Dataset details are summarized in Table~\ref{data}.

\begin{table}[ht]
\footnotesize

\caption{Statistics of Datasets. NC: Node Classification, LP: Link Prediction, CL: Node Clustering.}
\centering
\renewcommand{\arraystretch}{1}

\setlength{\tabcolsep}{2pt} 
\begin{tabular}{ccccc}
\hline
\label{data}
Dataset   & \# Nodes (\# Types)    & \# Edges  & Target (\# Labels)    & Task  \\ \hline
Yelp      & 31,081 (3)          & 411,263   & User-Business      & LP    \\  
Movielens & 2,672 (4)           & 234,695   & User-Movie         & LP    \\ \hline
DBLP      & 26,128 (4)          & 119,783   & Author (4)         & NC, CL   \\  
ACM       & 11,246 (3)          & 17,426    & Paper (3)          & NC, CL    \\  
IMDB      & 11,519 (3)          & 17,009    & Movie (3)          & NC, CL    \\  
Freebase  & 43,854 (4)          & 75,517    & Movie (3)          & NC, CL    \\ \hline
\end{tabular}

\end{table}

\noindent\textbf{Baselines.} We compare MF2Vec against SOTA heterogeneous graph neural networks (HGNNs), categorized as follows: (a) Relation-Based HGNNs, including RGCN, RGAT, HGT, HetSANN, and Simple-HGN; and (b) Metapath-Based HGNNs, comprising MP2Vec, HAN, MAGNN, HeCo, MHNF, HGMAE HMSG, and MECCH.

\noindent\textbf{Graph Construction Settings.}  
Following Algorithm 2, we ensured that the node types matched the target node and that the path length was at most 5. For each target node, we performed up to 1000 trials. The number of negative samples was set to 1.

\noindent\textbf{Experimental Settings.}  
To ensure a fair comparison among all HGNNs, the node embedding dimension was fixed to 64. The learning rate was set to \(1 \times 10^{-3}\), weight decay to \(1 \times 10^{-5}\), and the dropout rate to 0.5. For each baseline model, hyperparameters such as batch size, attention heads, dropout rate, learning rate, and epochs—excluding the node embedding dimension—followed the default settings specified in the original papers. Training was conducted using the Adam optimizer with early stopping (patience of 20 epochs) to prevent overfitting. Specifically, for MF2Vec, the temperature (\(\tau\)) was set to 0.5, and the number of facets (\(K\)) was set to 5. To demonstrate generalization, the experiments were conducted using five seed values (1, 10, 100, 1000, and 10000), with the data split into train (80\%), validation (10\%), and test (10\%). The model seeds were fixed during the experiments, and the average and standard deviation of the results were calculated for all seeds.

\begin{table*}[!ht]
\centering
\footnotesize

\setlength{\tabcolsep}{3pt} 
\renewcommand{\arraystretch}{1.15} 
\caption{Node classification performance. Best results are in bold, second-best are underlined. GraphHINGE does not support node classification.}

\begin{tabular}{l|l|ccc|ccc|ccc|ccc}
\hline
\label{table_1}
&Dataset   & \multicolumn{3}{c|}{DBLP}   & \multicolumn{3}{c|}{ACM}      & \multicolumn{3}{c|}{IMDB}   &\multicolumn{3}{c}{Freebase}    \\
\hline
&Model & AUC  &  {Mi-F1} &  {Ma-F1}    & AUC  &  {Mi-F1} &  {Ma-F1}   & AUC  &  {Mi-F1} &  {Ma-F1}  & AUC  &  {Mi-F1} &  {Ma-F1}    \\
\hline
\multirow{5}{*}{\rotatebox[origin=c]{90}{Relation-based}} 
&RGCN         &0.858{\tiny$\pm$0.004} &0.842{\tiny$\pm$0.005} &0.837{\tiny$\pm$0.005} &0.810{\tiny$\pm$0.012} &0.769{\tiny$\pm$0.007} &0.762{\tiny$\pm$0.011} &0.533{\tiny$\pm$0.024} &0.431{\tiny$\pm$0.027} &0.337{\tiny$\pm$0.043} &0.574{\tiny$\pm$0.049} &0.508{\tiny$\pm$0.057} &0.408{\tiny$\pm$0.078} \\
&RGAT         &0.849{\tiny$\pm$0.003} &0.829{\tiny$\pm$0.004} &0.823{\tiny$\pm$0.004} &0.803{\tiny$\pm$0.021} &0.761{\tiny$\pm$0.023} &0.749{\tiny$\pm$0.025} &0.558{\tiny$\pm$0.011} &0.424{\tiny$\pm$0.015} &0.409{\tiny$\pm$0.016} &0.622{\tiny$\pm$0.027} &0.553{\tiny$\pm$0.033} &0.487{\tiny$\pm$0.037} \\
&HGT          &0.926{\tiny$\pm$0.017} &0.889{\tiny$\pm$0.026} &0.899{\tiny$\pm$0.022} &0.779{\tiny$\pm$0.016} &0.741{\tiny$\pm$0.016} &0.709{\tiny$\pm$0.024} &0.540{\tiny$\pm$0.019} &0.396{\tiny$\pm$0.024} &0.386{\tiny$\pm$0.022} &0.578{\tiny$\pm$0.030} &0.482{\tiny$\pm$0.034} &0.427{\tiny$\pm$0.037} \\
&HetSANN      &0.950{\tiny$\pm$0.006} &0.931{\tiny$\pm$0.007} &0.925{\tiny$\pm$0.008} &0.780{\tiny$\pm$0.023} &0.736{\tiny$\pm$0.025} &0.711{\tiny$\pm$0.027} &0.613{\tiny$\pm$0.006} &0.487{\tiny$\pm$0.010} &0.484{\tiny$\pm$0.007} &0.646{\tiny$\pm$0.021} &0.575{\tiny$\pm$0.027} &0.521{\tiny$\pm$0.030} \\
&Simple-HGN   &0.947{\tiny$\pm$0.011} &0.928{\tiny$\pm$0.011} &0.920{\tiny$\pm$0.015} &0.840{\tiny$\pm$0.037} &0.808{\tiny$\pm$0.033} &\underline{0.793{\tiny$\pm$0.045}} &0.612{\tiny$\pm$0.035} &0.497{\tiny$\pm$0.038} &0.481{\tiny$\pm$0.045} &0.671{\tiny$\pm$0.014} &0.630{\tiny$\pm$0.017} &0.548{\tiny$\pm$0.031} \\
\hline
\multirow{8}{*}{\rotatebox[origin=c]{90}{Metapath-based}}
&MP2Vec       &0.586{\tiny$\pm$0.020} &0.399{\tiny$\pm$0.025} &0.372{\tiny$\pm$0.026} &0.731{\tiny$\pm$0.010} &0.663{\tiny$\pm$0.010} &0.633{\tiny$\pm$0.014} &0.511{\tiny$\pm$0.011} &0.397{\tiny$\pm$0.016} &0.378{\tiny$\pm$0.013} &0.605{\tiny$\pm$0.019} &0.603{\tiny$\pm$0.022} &0.506{\tiny$\pm$0.030} \\
&HAN          &0.937{\tiny$\pm$0.003} &0.919{\tiny$\pm$0.003} &0.909{\tiny$\pm$0.004} &0.785{\tiny$\pm$0.011} &0.762{\tiny$\pm$0.009} &0.729{\tiny$\pm$0.014} &0.550{\tiny$\pm$0.020} &0.426{\tiny$\pm$0.021} &0.384{\tiny$\pm$0.034} &0.665{\tiny$\pm$0.011} &0.643{\tiny$\pm$0.014} &0.517{\tiny$\pm$0.033} \\
&MAGNN        &0.972{\tiny$\pm$0.003} &0.926{\tiny$\pm$0.003} &0.915{\tiny$\pm$0.003} &\underline{0.945{\tiny$\pm$0.014}} &\underline{0.806{\tiny$\pm$0.031}} &0.777{\tiny$\pm$0.049} &\underline{0.739{\tiny$\pm$0.013}} &\underline{0.534{\tiny$\pm$0.032}} &0.492{\tiny$\pm$0.070} &0.744{\tiny$\pm$0.015} &0.674{\tiny$\pm$0.019} &0.518{\tiny$\pm$0.063} \\
&HeCo         &0.978{\tiny$\pm$0.003} &0.894{\tiny$\pm$0.013} &0.886{\tiny$\pm$0.014} &0.831{\tiny$\pm$0.019} &0.745{\tiny$\pm$0.012} &0.703{\tiny$\pm$0.013} &0.544{\tiny$\pm$0.006} &0.385{\tiny$\pm$0.014} &0.360{\tiny$\pm$0.004} &0.753{\tiny$\pm$0.021} &0.652{\tiny$\pm$0.021} &0.515{\tiny$\pm$0.033} \\
&MHNF         &0.925{\tiny$\pm$0.004} &0.896{\tiny$\pm$0.006} &0.886{\tiny$\pm$0.006} &0.883{\tiny$\pm$0.009} &0.760{\tiny$\pm$0.012} &0.711{\tiny$\pm$0.016} &0.530{\tiny$\pm$0.042} &0.408{\tiny$\pm$0.042} &0.324{\tiny$\pm$0.071} &0.512{\tiny$\pm$0.013} &0.451{\tiny$\pm$0.014} &0.265{\tiny$\pm$0.053} \\
&HGMAE        &0.948{\tiny$\pm$0.018} &0.874{\tiny$\pm$0.121} &0.829{\tiny$\pm$0.084} &0.812{\tiny$\pm$0.011} &0.732{\tiny$\pm$0.012} &0.736{\tiny$\pm$0.019} &0.614{\tiny$\pm$0.019} &0.427{\tiny$\pm$0.016} &0.353{\tiny$\pm$0.033} &0.727{\tiny$\pm$0.030} &0.622{\tiny$\pm$0.030} &0.478{\tiny$\pm$0.042} \\
&HMSG         &0.978{\tiny$\pm$0.003} &0.901{\tiny$\pm$0.020} &0.894{\tiny$\pm$0.021} &0.923{\tiny$\pm$0.013} &0.795{\tiny$\pm$0.019} &0.785{\tiny$\pm$0.020} &0.647{\tiny$\pm$0.021} &0.472{\tiny$\pm$0.017} &0.461{\tiny$\pm$0.022} &0.744{\tiny$\pm$0.026} &\underline{0.679{\tiny$\pm$0.040}} &0.531{\tiny$\pm$0.038} \\
&MECCH   & \underline{0.991{\tiny$\pm$0.133}} & \underline{0.941{\tiny$\pm$0.165}} &\underline{0.937{\tiny$\pm$0.176}} &0.916{\tiny$\pm$0.133} &0.790{\tiny$\pm$0.165} &\underline{0.793{\tiny$\pm$0.176}} &0.657{\tiny$\pm$0.133} &\underline{0.512{\tiny$\pm$0.165}}
&0.502{\tiny$\pm$0.176} & \underline{0.773{\tiny$\pm$0.133}} & 0.648{\tiny$\pm$0.165} & \underline{0.588{\tiny$\pm$0.176}} \\ \hline
\rowcolor[gray]{0.9} 
&\textbf{MF2Vec}   & \textbf{0.992{\tiny$\pm$0.002}} & \textbf{0.946{\tiny$\pm$0.011}} & \textbf{0.943{\tiny$\pm$0.010}} & \textbf{0.954}{\tiny$\pm$0.008} & \textbf{0.843}{\tiny$\pm$0.008} & \textbf{0.837}{\tiny$\pm$0.008} & \textbf{0.768}{\tiny$\pm$0.033} & \textbf{0.603}{\tiny$\pm$0.041} & \textbf{0.581}{\tiny$\pm$0.052}  & \textbf{0.828}{\tiny$\pm$0.023} & \textbf{0.694}{\tiny$\pm$0.021} & \textbf{0.629}{\tiny$\pm$0.047}
 \\ \hline
\end{tabular}
\vspace{-2mm}
\end{table*}

\subsection{Downstream Task}

\subsubsection{\textbf{Node Classification.}}
In our node classification experiments, we convert nodes into low-dimensional vector embeddings and apply a Softmax function to obtain probability distributions over label categories. We evaluate the model using Micro-F1, Macro-F1, and AUC metrics, providing a balanced assessment of classification performance. These metrics are used to compare our model, MF2Vec, against baseline models on the DBLP, ACM, IMDB and Freebase datasets, with results presented in Table \ref{table_1}. Our approach consistently outperforms baseline models across all metrics, demonstrating the effectiveness of MF2Vec. The significant performance improvement, particularly on the IMDB dataset, highlights the model's ability to capture complex relational data through multi-facet vectors. This suggests that MF2Vec effectively reveals nuanced relationships within heterogeneous information networks (HINs), enhancing understanding beyond node types.

\subsubsection{\textbf{Link Prediction.}}
In Movielens and Yelp, link prediction is framed as a binary classification task to determine the presence of an edge in the original graph. During training, validation, and testing, negative edges are created by replacing the destination node in positive edges with a randomly chosen node that is not connected to the source node. The link probability is computed using the dot product of node embeddings: $p_{v_{a},v_{b}} = \sigma(v_{a}^{T} \cdot v_{b})$, where $\sigma$ is the sigmoid function. Table \ref{link_prediction} presents the AUC, Micro-F1, and Macro-F1 scores for our model and baselines on Yelp and Movielens. MF2Vec performs competitively on Yelp (against MECCH) and outperforms all baselines on MovieLens. These results highlight the advantage of using multi-facet vectors derived from node paths for link prediction, and demonstrate the efficacy of our model in updating embeddings without relying on node types.

\begin{table}[htp]
\centering
\footnotesize

\caption{Link prediction performance. Best results are in bold, second-best are underlined. MHNF and HGMAE do not support link prediction.}

\label{link_prediction}
\setlength{\tabcolsep}{2pt} 
\renewcommand{\arraystretch}{1.15}
\begin{tabular}{l|ccc|ccc}
\hline
Dataset & \multicolumn{3}{c|}{Yelp}            & \multicolumn{3}{c}{Movielens}              \\ 
\hline
Model           & AUC    & Mi-F1    & Ma-F1    & AUC    & Mi-F1    & Ma-F1     \\ 
\hline
RGCN           & 0.635{\tiny$\pm$0.141} & 0.621{\tiny$\pm$0.103} & 0.614{\tiny$\pm$0.134} & 0.741{\tiny$\pm$0.003} & 0.664{\tiny$\pm$0.003} & 0.664{\tiny$\pm$0.003} \\
RGAT           & 0.697{\tiny$\pm$0.104} & 0.665{\tiny$\pm$0.081} & 0.655{\tiny$\pm$0.086}  & 0.687{\tiny$\pm$0.115} & 0.636{\tiny$\pm$0.101} & 0.630{\tiny$\pm$0.104} \\

HGT             & 0.832{\tiny$\pm$0.004} & 0.771{\tiny$\pm$0.004} & 0.770{\tiny$\pm$0.005}  & 0.755{\tiny$\pm$0.136} & 0.711{\tiny$\pm$0.116} & 0.711{\tiny$\pm$0.115} \\ 
HetSANN         & 0.703{\tiny$\pm$0.017} & 0.689{\tiny$\pm$0.013} & 0.688{\tiny$\pm$0.014} & 0.774{\tiny$\pm$0.013} & 0.772{\tiny$\pm$0.013} & 0.771{\tiny$\pm$0.013}  \\
Simple-HGN      & 0.824{\tiny$\pm$0.016} & 0.757{\tiny$\pm$0.018} & 0.757{\tiny$\pm$0.018}  & 0.820{\tiny$\pm$0.007} & 0.746{\tiny$\pm$0.012} & 0.745{\tiny$\pm$0.015} \\
 
\hline
MP2Vec         & 0.718{\tiny$\pm$0.003} & 0.658{\tiny$\pm$0.002} & 0.664{\tiny$\pm$0.003} & 0.798{\tiny$\pm$0.004} & 0.662{\tiny$\pm$0.004} & 0.669{\tiny$\pm$0.002} \\
HAN            & 0.627{\tiny$\pm$0.102} & 0.528{\tiny$\pm$0.076} & 0.399{\tiny$\pm$0.140} & 0.822{\tiny$\pm$0.004} & 0.642{\tiny$\pm$0.050} & 0.604{\tiny$\pm$0.072} \\ 
MAGNN          & 0.835{\tiny$\pm$0.005} & 0.773{\tiny$\pm$0.002} & 0.774{\tiny$\pm$0.003}  & \underline{0.858{\tiny$\pm$0.010}} & \underline{0.776{\tiny$\pm$0.008}} & \underline{0.781{\tiny$\pm$0.009}} \\ 

GraphHINGE & 0.839{\tiny$\pm$0.001} & 0.773{\tiny$\pm$0.001} & 0.766{\tiny$\pm$0.002} & 0.858{\tiny$\pm$0.004} & 0.775{\tiny$\pm$0.003} & 0.783{\tiny$\pm$0.004} \\

HeCo           & 0.838{\tiny$\pm$0.003} & 0.767{\tiny$\pm$0.004} & 0.755{\tiny$\pm$0.002} & 0.839{\tiny$\pm$0.003} & 0.755{\tiny$\pm$0.004} & 0.771{\tiny$\pm$0.005} \\

HMSG          & 0.851{\tiny$\pm$0.003} & \underline{0.784{\tiny$\pm$0.002}} & \underline{0.803{\tiny$\pm$0.004}} & 0.814{\tiny$\pm$0.006} & 0.744{\tiny$\pm$0.006} & 0.753{\tiny$\pm$0.013} \\ 
MECCH         & \underline{0.867{\tiny$\pm$0.012}} & 0.782{\tiny$\pm$0.013} & 0.774{\tiny$\pm$0.015} & 0.752{\tiny$\pm$0.006} & 0.690{\tiny$\pm$0.015} & 0.634{\tiny$\pm$0.040} \\ 
\hline
\rowcolor[gray]{0.9}
\textbf{MF2Vec} & \textbf{0.893{\tiny$\pm$0.003}} & \textbf{0.820{\tiny$\pm$0.005}} & \textbf{0.824{\tiny$\pm$0.004}} & \textbf{0.918{\tiny$\pm$0.003}} & \textbf{0.840{\tiny$\pm$0.004}} & \textbf{0.847{\tiny$\pm$0.007}} \\ 
\hline 
\end{tabular}

\end{table}

\subsubsection{\textbf{Node Clustering.}}
We carry out node clustering experiments using the DBLP, ACM, IMDB, and Freebase datasets to assess the embedding quality produced by MF2Vec. In these experiments, similar to node classification, we utilize the embedding vectors of target nodes from the testing set as input for the K-Means model and employ NMI and ARI metrics to gauge performance. As presented in Table \ref{table_3}, our model surpasses the baseline models, except for MHNF \cite{mhnf} in ARI on the IMDB dataset, where MF2Vec closely follows it by a small margin, demonstrating our model's superior ability to generate effective node representations based on multi-facet vectors for node clustering.

\begin{table}[!htp]
\centering

\caption{Node clustering. Best results are in bold, second-best are underlined. Std are omitted due to length constraints. GraphHINGE does not support node clustering.}
\footnotesize

\renewcommand{\arraystretch}{1}
\setlength{\tabcolsep}{3pt} 
\resizebox{\columnwidth}{!}{%
\begin{tabular}{l|cc|cc|cc|cc}
\hline
\label{table_3}

Dataset & \multicolumn{2}{c|}{DBLP}            & \multicolumn{2}{c|}{ACM}             & \multicolumn{2}{c|}{IMDB} & \multicolumn{2}{c}{Freebase}            \\ \hline
Model           & {NMI}    & ARI    & {NMI}    & ARI    & {NMI}    & ARI  & {NMI}    & ARI   \\ \hline 
RGCN       &  0.0168 & 0.0670 & 0.3612 & 0.2809 & 0.0088 & 0.0056 & 0.0228 & 0.0443\\
RGAT       & 0.4435 & 0.4097 & 0.3976 & 0.3140 & 0.0477 & 0.0143 & 0.0327 & 0.0387\\
HGT        & {0.7722} & 0.8250 & {0.5064} & 0.5335 & {0.0433} & {0.0459} &0.0801 &0.1075\\ 
HetSANN    &0.752 &0.7723 & 0.5274 & 0.5263 & 0.0865 & 0.043 & 0.1054 & 0.0967 \\
Simple-HGN &0.7707 & 0.6405 &0.5309 &0.5237 & 0.1410 &0.1180 &0.1541 & 0.1804\\ \hline
MP2Vec     & {0.0519} & 0.0330  & {0.1034} & 0.0641 & {0.0065} & 0.0096 &0.0145 & 0.0286 \\ 
HAN        & {0.3533} & 0.1845 & {0.3036} & 0.3528 & {0.0057} & 0.0015 & 0.0211 & 0.0217 \\ 
MAGNN      & {0.7964} & 0.8193 & {0.5125} & 0.5317 & {0.1075} & {0.1199} &0.1872 &0.1477\\ 

HeCo       & {0.7477} & 0.7872 & \underline{0.5394} & \underline{0.5454} & {0.0049} & 0.0028 &0.0805 & 0.0793 \\ 
MHNF       & {0.7197} & 0.7498 & 0.5385 & 0.5274 & \underline{0.1443} & \textbf{0.1467 }  & 0.1881 & 0.1697   \\ 
HGMAE       & {0.7076} & 0.7375 & 0.5315 & 0.5321 & {0.1264} & {0.1345 }  & 0.2401 & 0.2485   \\ 
HMSG       & {0.7278} & 0.7767 & {0.4475} & 0.3630  & {0.0335} & 0.0326 &0.1257 &0.1213 \\ 

MECCH      & \underline{0.8312} & \underline{0.8739} & {0.4202} & 0.3513 & {0.0070}  & 0.0056 & \underline{0.2612} & \underline{0.3127}\\ \hline 
\rowcolor[gray]{0.9}
\textbf{MF2Vec}       & \textbf{{0.8606}} & \textbf{0.8951} & \textbf{{0.5714}}  & \textbf{0.5790} & \textbf{{0.1459}} & \underline{0.1448}  & \textbf{0.3083} & \textbf{0.3729}\\ \hline
\end{tabular}
}

\end{table}

\subsection{Stability under Meta-path Variations}

This section empirically evaluates whether MF2Vec can achieve consistent and robust performance across various settings by effectively leveraging multi-facet information shared among all nodes, regardless of the presence or absence of node type-based metapaths. All experiments are conducted on three datasets, and the relative advantages of MF2Vec are assessed through comparison with representative metapath-based GNNs (MAGNN, HMSG, MECCH). The details and interpretations of each experiment are as follows.

\subsubsection{\textbf{Consistency Across Metapath Types}}
Table~\ref{std} presents the performance variability of each model when trained and evaluated on different metapath types (e.g., APA, APCPA, etc.). For each dataset, we measure the standard deviation (Std) of AUC, Mi-F1, and Ma-F1 scores, and find that MF2Vec consistently achieves the lowest standard deviation, indicating the highest stability. This result suggests that MF2Vec does not rely on specific metapaths defined by node types, but instead learns consistent and reliable representations by leveraging shared facet information among nodes.

\begin{table}[ht]
\scriptsize

\caption{Standard deviation (Std) of node classification performance for single metapaths across datasets. Lower values indicate greater stability.}
\label{std}

\renewcommand{\arraystretch}{1}
\centering
\begin{tabular}{c|c|ccc|>{\columncolor[gray]{0.9}}c}
\hline
Dataset & Metric & MAGNN  & HMSG   & MECCH  & \textbf{MF2Vec} \\ \hline
\multirow{3}{*}{DBLP}  
&AUC     & 0.130 & 0.177 & \underline{0.107} & \textbf{0.060}\\
&Mi-F1   & 0.217 & 0.222 & \underline{0.173} & \textbf{0.127} \\ 
&Ma-F1   & 0.201 & 0.257 & \underline{0.170} & \textbf{0.125} \\\hline
\multirow{3}{*}{ACM}                       
&AUC     & 0.038 & 0.039 & \underline{0.034} & \textbf{0.032} \\
&Mi-F1   & 0.078 & 0.058 & \underline{0.051} & \textbf{0.033} \\ 
&Ma-F1   & \underline{0.052}& 0.075 & 0.062 & \textbf{0.049} \\ \hline
\multirow{3}{*}{IMDB}                    
&AUC     & \underline{0.034} & 0.039 & 0.041 & \textbf{0.025} \\
&Mi-F1   & \underline{0.020} & 0.023 & 0.049 & \textbf{0.008} \\ 
&Ma-F1   & \underline{0.015} & 0.046 & 0.079 & \textbf{0.010} \\ \hline
\end{tabular}

\end{table}

\subsubsection{\textbf{Generalization Across Unseen Metapaths}}
\begin{figure}[ht]
    \centering
    \includegraphics[width=0.9\linewidth]{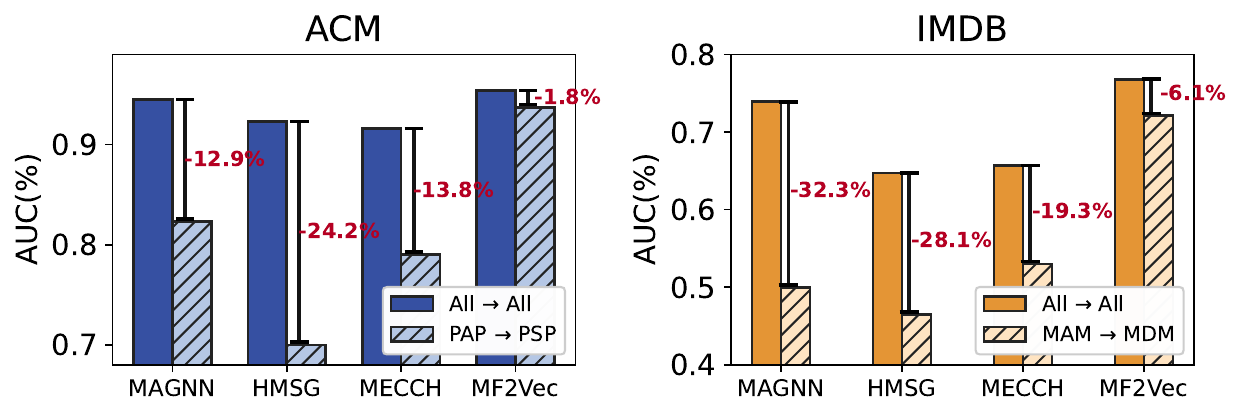}
    \Description{Comparison of cross-metapath generalization performance across models. MF2Vec shows smallest performance drop.}
    \caption{
    $\rightarrow$ indicates (train path $\rightarrow$ test path). The red percentage shows the performance drop when generalizing from all-path training/testing to cross-metapath settings.
    }
    \label{fig:cross_metapath}
\end{figure}

Figure~\ref{fig:cross_metapath} compares the generalization performance of each model when trained on a single metapath and tested on different, unseen metapaths. MF2Vec consistently achieves the highest generalization performance even under these unseen conditions, and the performance drop compared to using all metapaths together is significantly smaller than that of other models. This indicates that MF2Vec is less dependent on specific metapath schemas and is able to extract more transferable and robust representations. These results confirm that the ability to select facets independently of explicit path structures is crucial for achieving stable and generalized performance.

\begin{table*}[ht]
\centering
\caption{Performance comparison of different methods for reflecting edge weights across various datasets. No.: No edge weight; Random.: Random selection; STE: Straight-Through Estimator~\cite{bengio2013estimating}; Gumble.: Gumble Softmax. The best results are highlighted in bold, and the second-best results are underlined.
}

\label{table8}
\setlength{\tabcolsep}{1pt} 
\resizebox{\textwidth}{!}{%
\begin{tabular}{@{}l|ccc|ccc|ccc|ccc|ccc|ccc@{}}
\toprule
Dataset               & \multicolumn{3}{c|}{Yelp}                      & \multicolumn{3}{c|}{Movielens}                & \multicolumn{3}{c|}{DBLP}                     & \multicolumn{3}{c|}{ACM}                      & \multicolumn{3}{c|}{IMDB}                     & \multicolumn{3}{c}{Freebase}                \\  \hline
Method                    & AUC           & Mi-F1         & Ma-F1         & AUC           & Mi-F1         & Ma-F1         & AUC           & Mi-F1         & Ma-F1         & AUC           & Mi-F1         & Ma-F1         & AUC           & Mi-F1         & Ma-F1         & AUC           & Mi-F1         & Ma-F1         \\ \hline
No.    & 0.813{\tiny$\pm$0.003} & 0.743{\tiny$\pm$0.002} & 0.752{\tiny$\pm$0.004} 
       & 0.832{\tiny$\pm$0.003} & 0.754{\tiny$\pm$0.005} & 0.743{\tiny$\pm$0.013}    & 0.974{\tiny$\pm$0.004} & 0.868{\tiny$\pm$0.012} & 0.854{\tiny$\pm$0.012}
 & 0.912{\tiny$\pm$0.014} & 0.788{\tiny$\pm$0.015} & 0.759{\tiny$\pm$0.025}
       & 0.531{\tiny$\pm$ 0.022} & 0.378{\tiny$\pm$ 0.005} & 0.232{\tiny$\pm$ 0.036}     & 0.731{\tiny$\pm$0.011} & 0.646{\tiny$\pm$0.021} & 0.475{\tiny$\pm$0.015}
       \\
Random.& 0.865{\tiny$\pm$0.008} & 0.792{\tiny$\pm$0.004} & 0.803{\tiny$\pm$0.009} & 
0.869{\tiny$\pm$0.006} & 0.793{\tiny$\pm$0.007} & 0.802{\tiny$\pm$0.004} & 
0.985{\tiny$\pm$0.004} & 0.932{\tiny$\pm$0.007} & 0.930{\tiny$\pm$0.005} &
0.945{\tiny$\pm$0.005} & 0.823{\tiny$\pm$0.008} & 0.812{\tiny$\pm$0.006} & 
0.748{\tiny$\pm$0.008} & 0.584{\tiny$\pm$0.003} & 0.569{\tiny$\pm$0.007} & 
0.824{\tiny$\pm$0.008} & 0.691{\tiny$\pm$0.008} & 0.626{\tiny$\pm$0.007}

\\
STE.&\underline{0.867}{\tiny$\pm$0.018} & \underline{0.797}{\tiny$\pm$0.003} & \underline{0.808}{\tiny$\pm$0.015} & \underline{0.871}{\tiny$\pm$0.006} & \underline{0.795}{\tiny$\pm$0.006} & \underline{0.803}{\tiny$\pm$0.005} & \underline{0.989}{\tiny$\pm$0.003} & \underline{0.935}{\tiny$\pm$0.009} & \underline{0.931}{\tiny$\pm$0.01} 
& \underline{0.949}{\tiny$\pm$0.009} & \underline{0.824}{\tiny$\pm$0.013} & \underline{0.813}{\tiny$\pm$0.012} & \underline{0.751}{\tiny$\pm$0.005} & \underline{0.588}{\tiny$\pm$0.004} & \underline{0.571}{\tiny$\pm$0.003} & \underline{0.825}{\tiny$\pm$0.020} & \textbf{0.696}{\tiny$\pm$0.017} & \underline{0.628}{\tiny$\pm$0.042}
\\
\rowcolor[gray]{0.9}
Gumble.        & \textbf{0.893}{\tiny$\pm$0.003} & \textbf{0.820}{\tiny$\pm$0.005} & \textbf{0.824}{\tiny$\pm$0.004} & \textbf{0.918}{\tiny$\pm$0.003} & \textbf{0.840}{\tiny$\pm$0.004} & \textbf{0.847}{\tiny$\pm$0.007} & \textbf{0.991}{\tiny$\pm$0.002} & \textbf{0.946}{\tiny$\pm$0.011} & \textbf{0.943}{\tiny$\pm$0.010} & \textbf{0.954}{\tiny$\pm$0.008} & \textbf{0.843}{\tiny$\pm$0.008} & \textbf{0.837}{\tiny$\pm$0.008} & \textbf{0.768}{\tiny$\pm$0.033} & \textbf{0.603}{\tiny$\pm$0.041} & \textbf{0.581}{\tiny$\pm$0.052}  & \textbf{0.828}{\tiny$\pm$0.023} & \underline{0.694}{\tiny$\pm$0.021} & \textbf{0.629}{\tiny$\pm$0.047}
 \\ \bottomrule
\end{tabular}%
}
\end{table*}

\subsection{Time Complexity}
\begin{table}[H]
\centering
\scriptsize
\caption{Comparison of Time Complexity. $M$: Number of metapaths, $P$: Metapath length, $t^P$: Growth of neighbors with $P$, $d_{\text{edge}}$: Edge embedding dimension, $H$: Attention heads.}
\label{tab:time_complexity_comparison}
\begin{tabular}{l|l}
\hline
 {Model}          &  {Time Complexity}                     \\ \hline
 {HetSANN}        & $O(N \times H \times t \times d) + O(N \times H \times (t + d))$ \\
 {SimpleHGN}      & $O(N \times H \times t \times d) + O(N \times t \times d_{\text{edge}})$ \\ \hline
 {MAGNN}          & $O(N \times M \times t^P \times d \times P)$  \\ 
 {HAN}            & $O(N \times M \times t^P \times d)$           \\ 
 {MECCH}          & $O(N \times M \times t^P \times d)$           \\ \hline
 \rowcolor[gray]{0.9}
\textbf{MF2Vec}   & \textbf{\(O(N \times K \times t \times d)\)} \\ \hline
\end{tabular}
\end{table}
The proposed model's time complexity is $O(N \times K \times t \times d)$, where $N$ is the number of nodes, $t$ is the average neighbors, $K$ is the number of facets, and $d$ is the embedding dimension. Attention adds $O(N \times t \times K \times d)$, and graph convolution contributes $O(N \times t \times d)$, resulting in a total complexity of $O(N \times K \times t \times d)$. In Table~\ref{tab:time_complexity_comparison}, models like MECCH \cite{mecch}, HAN \cite{han}, and MAGNN \cite{magnn} rely on $M$ metapaths, incurring higher costs due to $t^P$ growth with metapath length ($P$). MF2Vec avoids this by leveraging facets ($K$), leading to more efficient computation, satisfying $MF2Vec < MECCH \approx HAN < MAGNN$.Additionally, we compare MF2Vec with relation-based methods like HetSANN~\cite{HetSANN} and SimpleHGN~\cite{SimpleHGN}, which also avoid metapaths by focusing on edge types.

\begin{figure}[ht]
    \centering
    \includegraphics[width=0.9\linewidth]{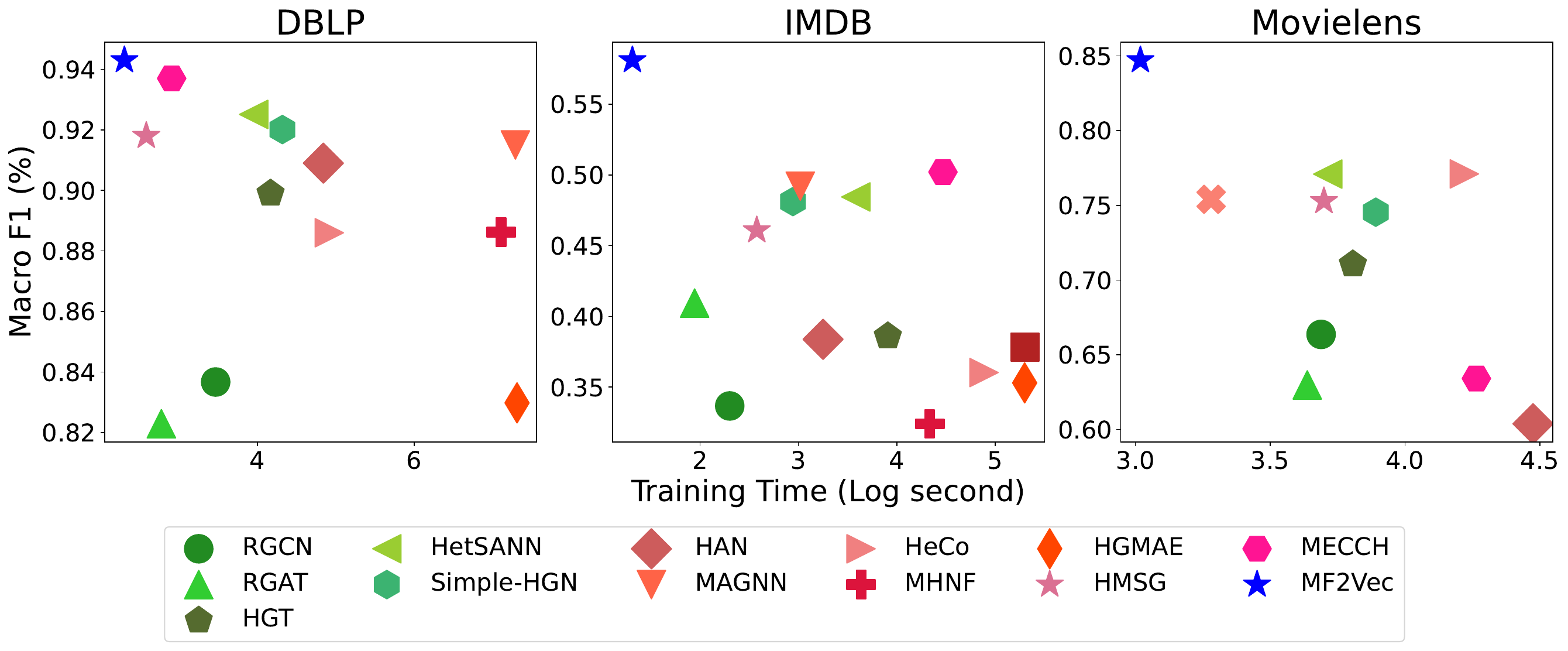}
    \Description{Training time (log seconds) vs Macro F1 across three datasets. MF2Vec achieves best tradeoff with faster convergence.}
    \caption{Training time (log seconds) and Macro F1 score across three datasets. Training time is measured until early stopping. The blue star (\textcolor{blue}{$\star$}) indicates MF2Vec.}
    \label{time_complexity}
\end{figure}

Figure \ref{time_complexity} compares training times and Macro F1 scores for MF2Vec and other models on DBLP, IMDB, and Movielens. MAGNN \cite{magnn} and MP2Vec \cite{metapath2vec} are excluded from Movielens due to their excessive training times. MF2Vec achieves the best performance on node classification and link prediction tasks, with significantly lower training times. This efficiency arises from its dynamic use of path characteristics, avoiding the high costs of aggregating multiple paths as seen in MAGNN and MP2Vec, ensuring better scalability and faster convergence without compromising performance.

%% file: 6.Ablation.tex
\subsection{Effect of Facet Selection Strategy}
Since MF2Vec selects the optimal facet for each path in a learnable manner, we conducted an additional experiment to validate the effectiveness of its facet selection strategy. Table \ref{table8} compares various methods for selecting facet information as edge weights. Our GCN-based results show that the proposed method, which leverages Gumbel-Softmax and attention mechanisms, consistently outperforms all alternatives across datasets. When using random facet selection, performance decreases, and omitting facet information entirely leads to the lowest results. We also evaluated the STE (Straight-Through Estimator)-based approach \cite{bengio2013estimating}, which enables probabilistic gradient propagation during facet selection. However, this method performs slightly worse than our proposed approach. These findings highlight that facet-related information is crucial for effective graph learning, and our method excels at capturing complex structures and hidden relationships.

\subsection{Visualization}

\begin{figure}[ht]
    \centering
    \includegraphics[width=0.8\linewidth]{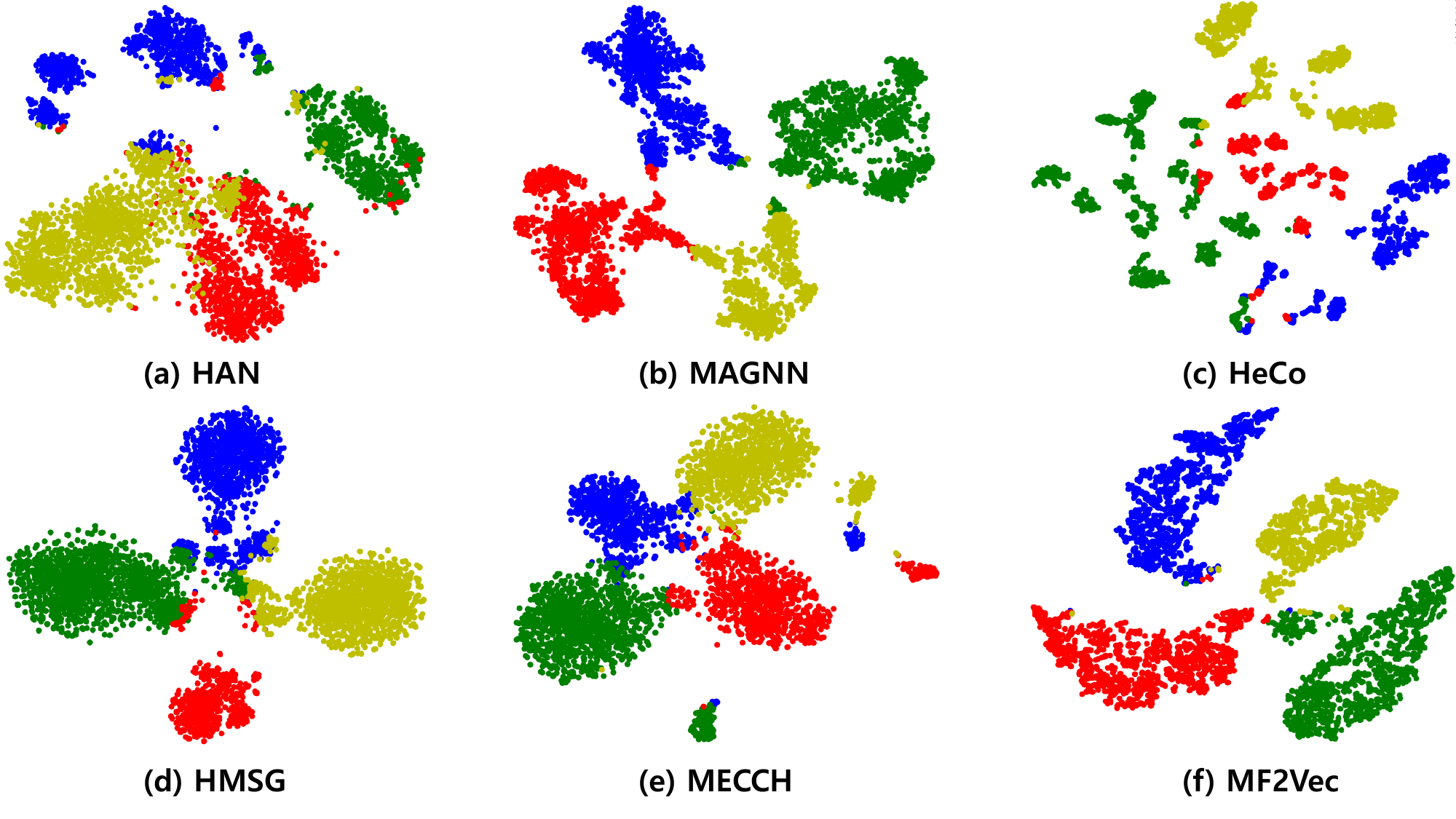}
    \caption{Visualization of the node embedding on DBLP.}
    \Description{Scatter plot showing the learned embeddings of nodes in DBLP, colored by label, demonstrating clear clustering by class.}
    \label{fig_3}
\end{figure}

To evaluate embedding quality, we visualize node representations from various HGNNs using t-SNE \cite{t-SNE}. Figure \ref{fig_3} shows t-SNE plots for HAN \cite{han}, MAGNN \cite{magnn}, HeCo \cite{wang2021self}, HMSG \cite{hmsg}, MECCH \cite{mecch}, and MF2Vec, with colors representing author classes. MF2Vec demonstrates superior clustering within classes and clear decision boundaries, outperforming the more scattered intra-class distributions and weaker inter-class separations of other HGNNs.

\subsection{Sensitivity to Random Walk}

\subsubsection{\textbf{Effect of Random Walk Length.}}
In Table~\ref{length}, increasing the random walk length potentially introduces more noise into the sampled paths, but the performance of MF2Vec remains stable across all tested values. Changes in AUC are minor, demonstrating that MF2Vec is insensitive to longer or noisier random walks. Based on these results, a length of 5 was used for experiments.

\begin{table}[h]
\centering
\footnotesize
\vspace{-3mm}
\caption{AUC of MF2Vec for varying random walk lengths. $\Delta$ is the max-min AUC per dataset.}
\vspace{-3mm}
\label{length}
\begin{tabular}{c|cccc|c}
\toprule
Dataset & 3 & 5 & 7 & 9 & $\Delta$ \\
\midrule
DBLP & 0.990 {\tiny$\pm$0.008} & 0.992 {\tiny$\pm$0.002} & 0.988 {\tiny$\pm$0.010} & 0.982 {\tiny$\pm$0.009} & 0.010 \\
ACM  & 0.939 {\tiny$\pm$0.010} & 0.954 {\tiny$\pm$0.008} & 0.953 {\tiny$\pm$0.007} & 0.951 {\tiny$\pm$0.011} & 0.015 \\
IMDB & 0.748 {\tiny$\pm$0.025} & 0.768 {\tiny$\pm$0.033} & 0.772 {\tiny$\pm$0.021} & 0.779 {\tiny$\pm$0.014} & 0.031 \\
\bottomrule
\end{tabular}
\vspace{-3mm}
\end{table}

\subsubsection{\textbf{Effect of Random Walk Quality.}}
\begin{figure}[ht]
    \centering
    \includegraphics[width=\linewidth]{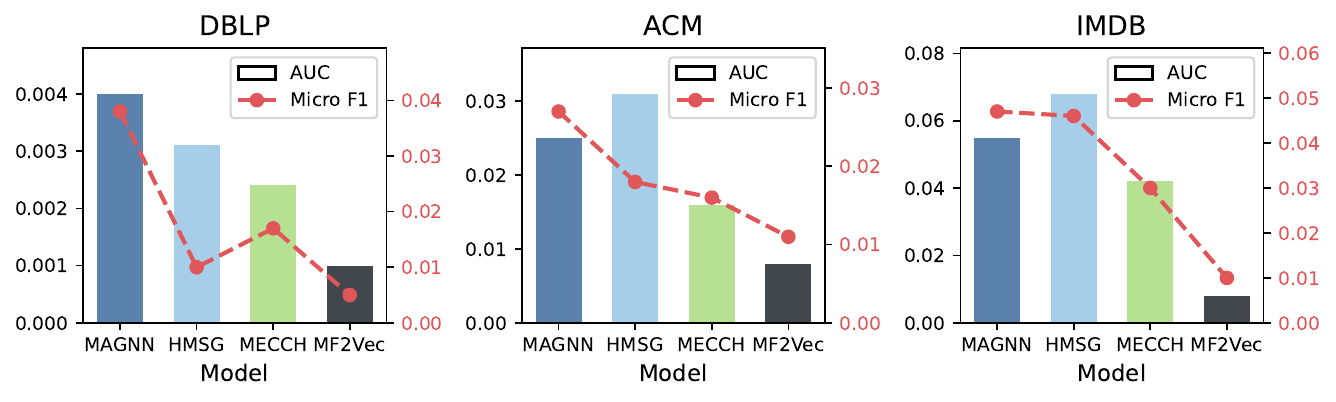}
    \caption{Robustness of MF2Vec and baseline models under varying edge sparsity (0\%, 10\%, 20\% edge removal).}
    \Description{Line chart comparing robustness of MF2Vec and other models when 0, 10, or 20 percent of edges are randomly removed. MF2Vec shows consistently higher performance across settings.}
    \label{fig:enter-label}
\end{figure}

We simulate sparse environments by randomly removing 0\%, 10\% and 20\% of node edges, then evaluate AUC and its standard deviation. As shown in Figure~\ref{fig:enter-label}, MF2Vec shows the least performance variation, maintaining robust results despite increased noise and missing data. This demonstrates that MF2Vec leverages shared facet information to generalize well in incomplete or noisy settings.

%% file: 8.Casestudy.tex
\subsection{Analysis of Facet Balance}

\begin{figure}[ht]
\centering
\includegraphics[width=0.8\linewidth]{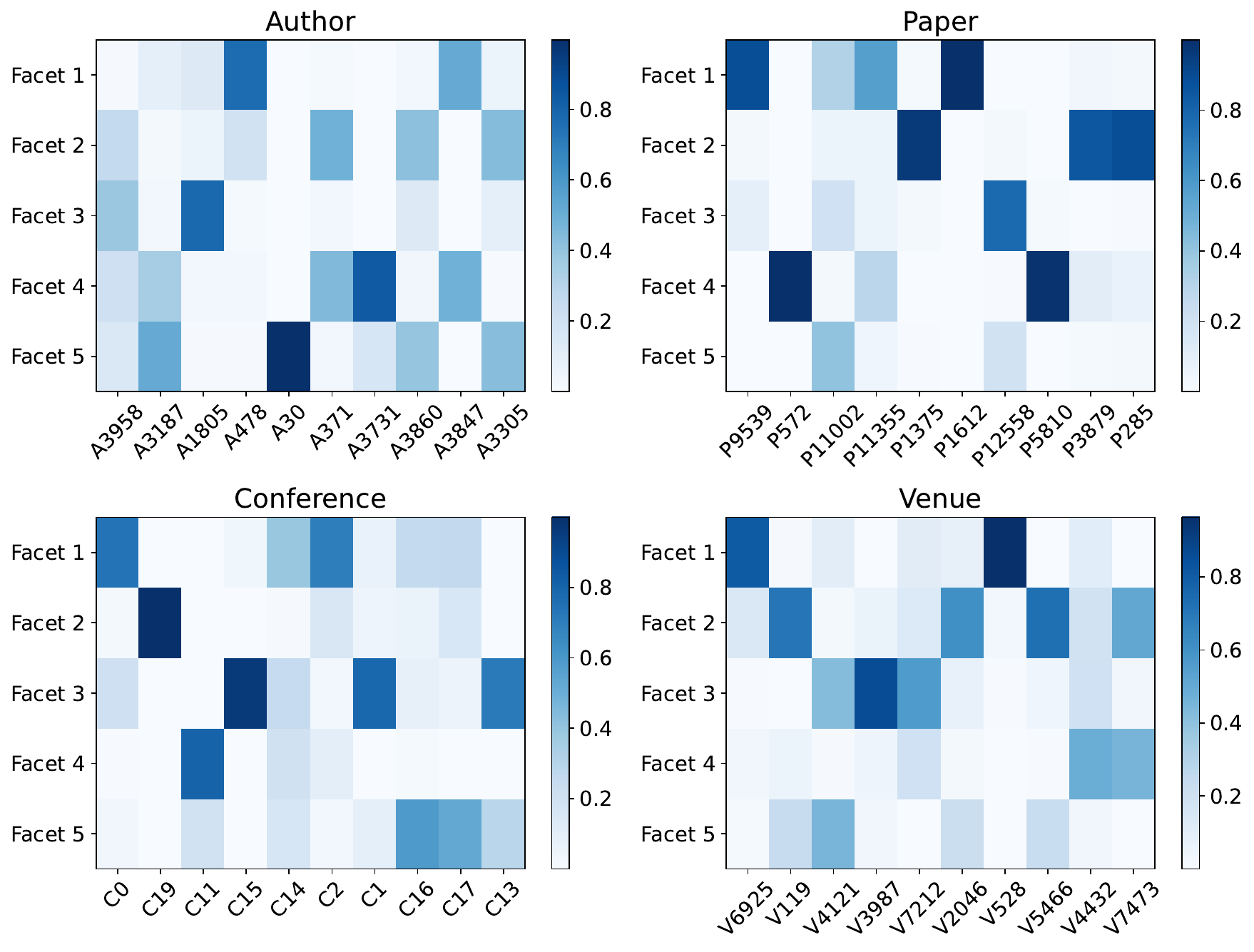}
\caption{Facet-level attention heatmap for the DBLP dataset. Columns represent 10 sampled nodes per type, and rows indicate shared facets across node types. Darker colors denote facets with greater significance in node connections.}
\Description{A heatmap showing facet-level attention across nodes in the DBLP dataset. Columns correspond to sampled nodes of different types, rows correspond to facets, and darker cells represent more influential facets.}
\label{case1}
\end{figure}

To further analyze the balance of facet distributions across node types, we visualized a facet-level attention heatmap as shown in Figure~\ref{case1}. Each column represents a specific node and each row corresponds to a facet. Darker colors indicate facets that play a more significant role in the node’s connections. For example, the paper node “P1612” mainly focuses on Facet 1, whereas the author node “A3958” distributes attention relatively evenly across all facets. This analysis confirms that some nodes concentrate on specific facets while others distribute attention more uniformly, with no strong bias overall. This visually validates the importance of modeling shared facets in heterogeneous graphs.

\subsection{Case Study}

\begin{figure}[ht]
\centering
\includegraphics[width=0.8\linewidth, trim=0 300 280 0, clip]{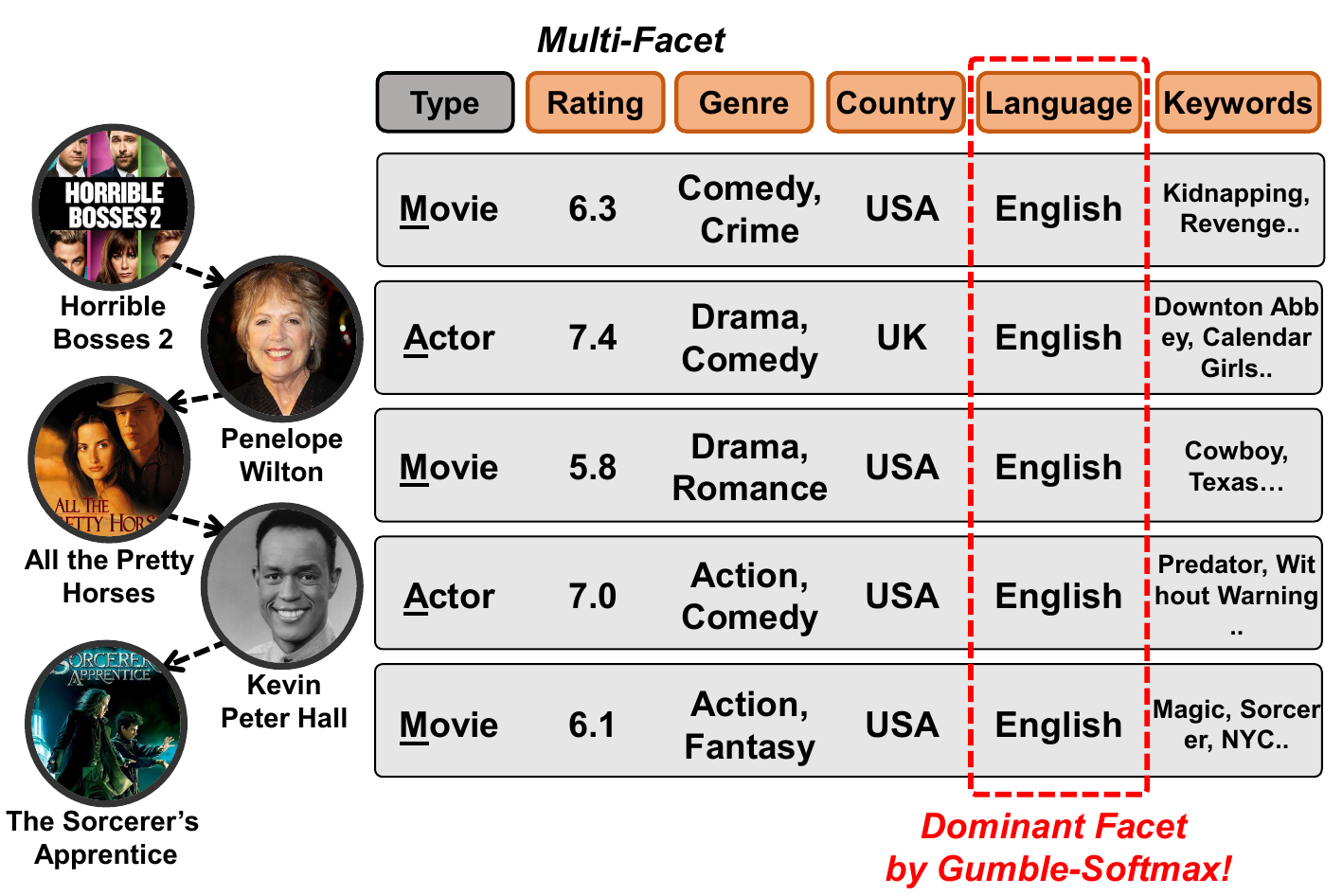}
\caption{Example path in the IMDB graph where each node is annotated with key attributes. All nodes along the path share “language: English” as the dominant facet.}
\Description{A path in the IMDB graph with nodes annotated by attributes such as rating, country, and language. All nodes share the language English, showing semantic grouping by facet.}
\label{case2}
\end{figure}

Most graph datasets represent nodes by unique IDs, making it difficult to interpret learned embeddings in terms of real-world attributes. To address this, we conducted a qualitative case study to investigate whether the learned facets actually correspond to meaningful attributes. Based on the IMDB text data provided in MAGNN, we used SBERT to generate node embeddings and separated them into five explicit attributes shared by IMDB node types: rating, genre, country, language, and keywords.
In this setting, we analyzed which attributes were shared among the intermediate nodes along a given path from a start node to an end node. For example, along the path “Horrible Bosses 2 → Penelope Wilton → All the Pretty Horses → Kevin Peter Hall → The Sorcerer’s Apprentice,” all nodes shared the attributes country (USA, UK), rating (around 6), and language (English). Notably, Gumbel-Softmax identified “language (English)” as the dominant facet. This shows that the connections in this path are not just simple Movie–Actor–Movie–Actor–Movie links, but are semantically grouped by language, thus demonstrating that MF2Vec learns interpretable and meaningful facets rather than mere latent ID embeddings.

\subsection{Ablation Study}

\subsubsection{\textbf{Effect of Temperature ($\tau$)}}

\begin{figure}[ht]
    \centering
    \includegraphics[width=0.8\linewidth]{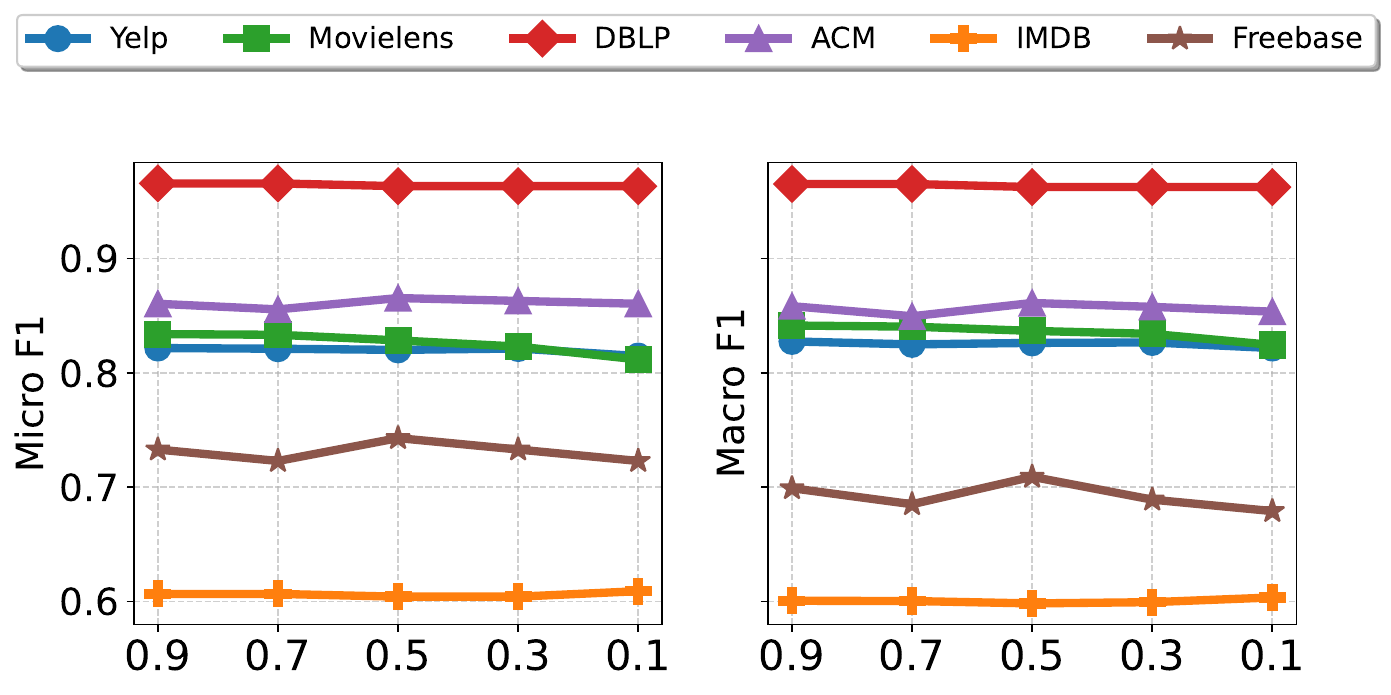}
    \caption{Performance of MF2Vec on six datasets with varying Temperature ($\tau$).}
    \Description{A line plot showing MF2Vec performance across six datasets at different temperature values in Gumbel-Softmax. Performance varies slightly but remains stable overall.}
    \label{tau_plot}
\end{figure}

From Figure \ref{tau_plot}, we observe the sensitivity of the Gumbel-Softmax in calculating the probability across all classes with varying $\tau$. According to the performance metrics, the model's performance exhibits minor variations in Movielens, DBLP, and Yelp, and changes within a 2\% range in Movielens and Freebase. The results indicate that the sensitivity of the model to different values of $\tau$ is relatively stable across datasets, with minor fluctuations in the Macro F1 and Micro F1 scores.

\subsubsection{\textbf{Effect of Train Ratio Settings.}}

\begin{figure}[ht]
    \centering
    \includegraphics[width=1\linewidth]{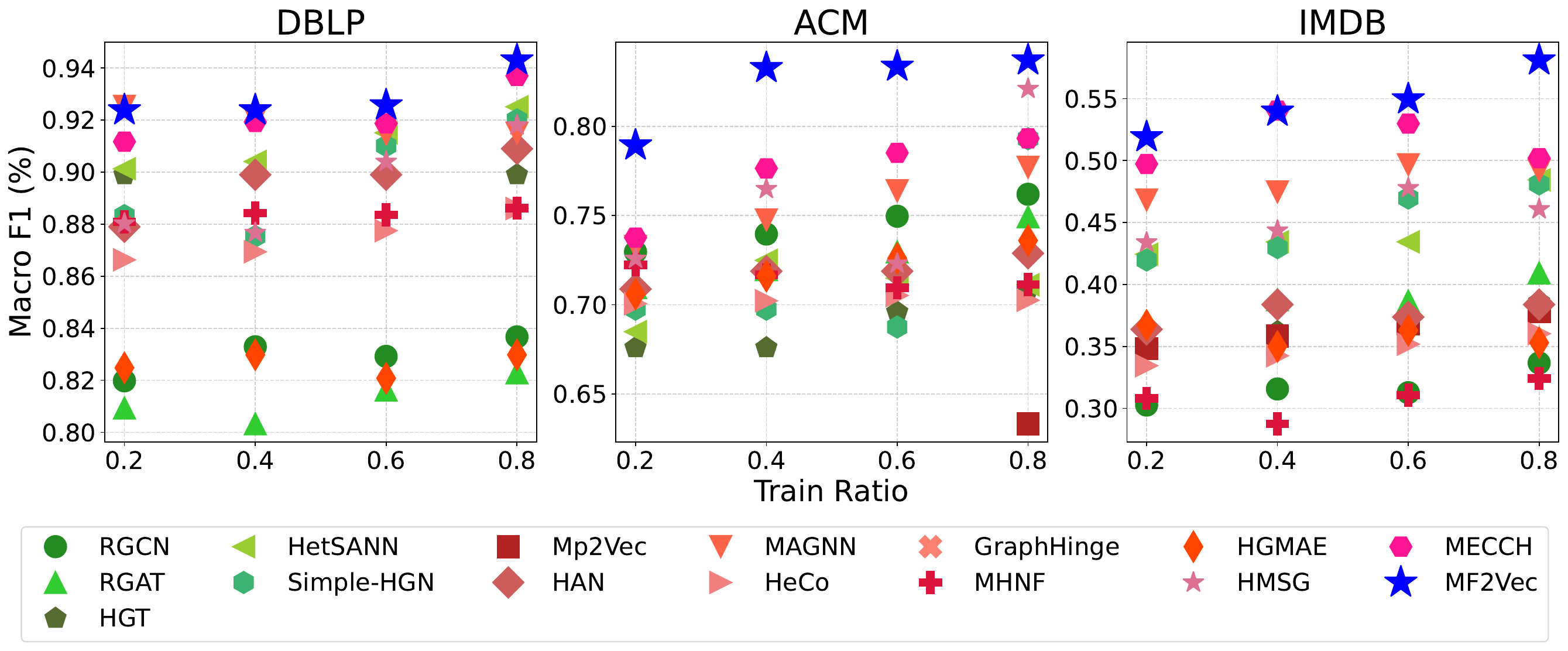}
    \caption{Performance of MF2Vec on three datasets with varying training set ratios, ranging from 0.8 to 0.2, in the node classification task.}
    \Description{A line chart showing classification performance of MF2Vec compared to baselines when varying the training set ratio from 0.8 to 0.2. MF2Vec consistently outperforms baselines.}
    \label{train_plot}
\end{figure}

As shown in Figure \ref{train_plot}, MF2Vec consistently achieves the highest performance across varying training set ratios, demonstrating its robustness and adaptability. While the performance of all models declines as the training ratio decreases, MF2Vec remains superior in most cases and retains competitiveness in the rest. At the lowest ratio of 0.2, it occasionally shows comparable performance to other models. Only models achieving scores above a certain threshold on each dataset are displayed to ensure clearer visualization. These results highlight MF2Vec's effectiveness in handling sparse training data.

\subsubsection{\textbf{Effect of Number of Facets.}}

\begin{figure}[ht]
    \centering
    \includegraphics[width=0.8\linewidth]{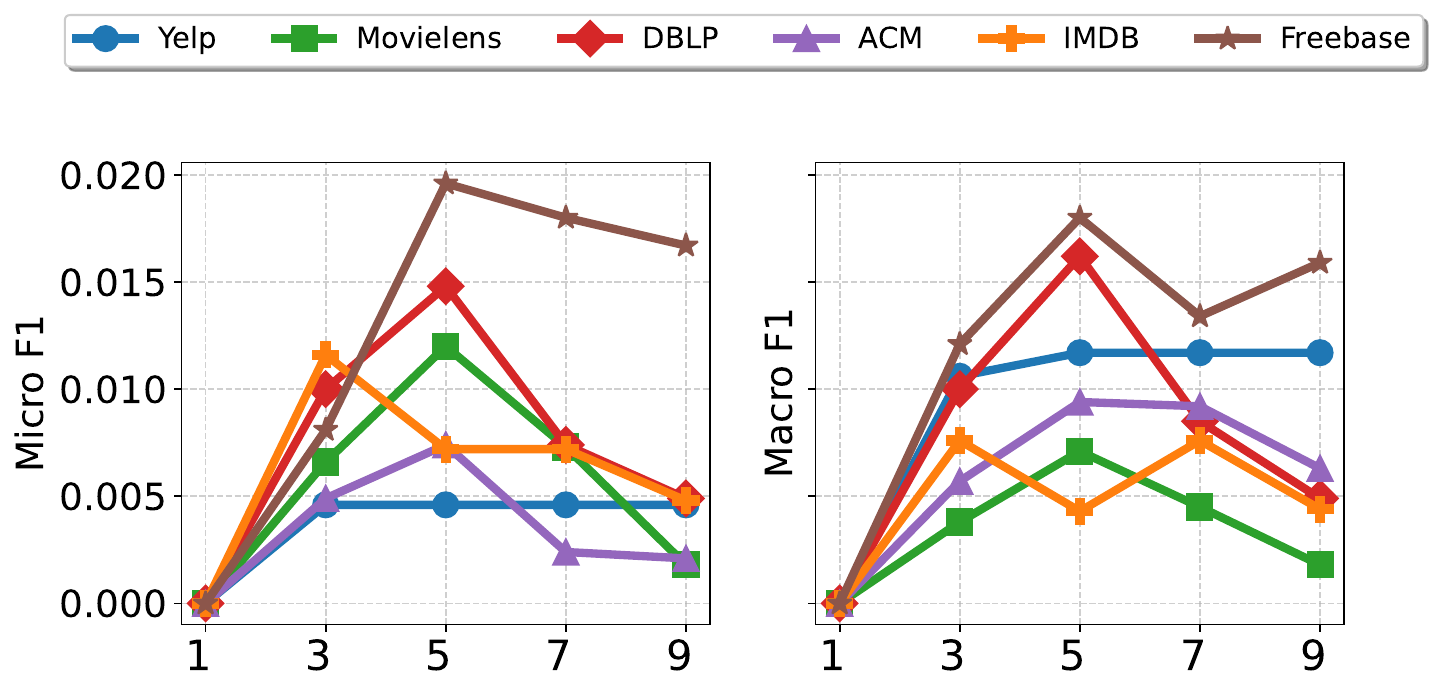}
    \caption{Performance of MF2Vec on six datasets with varying number of facets extracted from a path.}
    \Description{A plot showing MF2Vec's performance trend as the number of facets increases. Best performance occurs at five facets, with degradation beyond that point.}
    \label{facet_plot}
\end{figure}

Figure \ref{facet_plot} shows MF2Vec's performance across six datasets with varying facet counts. The performance difference relative to ``$\mathcal{K} = 1$" highlights the advantage of multiple facets. Except for IMDB, optimal results are achieved with five facets, consistently outperforming the single-facet case by effectively leveraging node and path information. Beyond five facets, performance declines, likely due to reduced feature grouping efficiency.

%% file: 2.Relatedwork_short.tex
\section{Related Work} 

\subsection{Heterogeneous Graph Neural Network}
\label{related_work}

Graph neural networks (GNNs) effectively handle graph-structured data by learning low-dimensional node representations \cite{defferrard, kipf, hamilton, velickovic}. Building on GNNs, heterogeneous graph neural networks (HGNNs) extend representation learning to heterogeneous graphs, enabling tasks like node classification, link prediction, and clustering. Message-passing methods have become prominent, generating embeddings by aggregating neighbor information.

\subsubsection{Relation-based HGNNs}

Relation-based HGNNs aggregate neighbor information with type-specific weights, eliminating manual meta-path definitions. RGCN \cite{chen2019rgcn} applies relation-specific GCNs, while RGAT \cite{RGAT} integrates self-attention for relational interactions. HetSANN \cite{HetSANN} and HGT \cite{hu2020heterogeneous} use meta-relation-based attention, and Simple-HGN \cite{SimpleHGN} combines type-level and node-level attention with L2 normalization. While efficient, deeper layers may cause over-smoothing \cite{LiQ} and performance degradation.

\subsubsection{Metapath-based HGNNs}

Metapath-based HGNNs analyze complex networks using predefined metapaths, enabling direct links between homogeneous nodes. Metapath2vec \cite{metapath2vec} uses random walks with a skip-gram model \cite{skip-gram}, while HAN \cite{han} and MAGNN \cite{magnn} aggregate features along metapaths. GTN \cite{GTN} dynamically learns metapaths, and GraphHINGE \cite{graphhinge} efficiently aggregates interactions. Models like HeCo \cite{wang2021self}, HMSG \cite{hmsg}, MHNF \cite{mhnf}, and MECCH \cite{mecch} enhance robustness and reduce redundancy. Despite progress, most approaches rely on single-facet paths, limiting their granularity. To address this, MF2Vec introduces a multi-faceted, fine-grained approach for richer node interactions and network understanding.

\subsection{Multiple-Vector Network Embedding}

Recent studies in graph representation learning focus on learning multiple embeddings for nodes, capturing their multi-faceted nature. PolyDW \cite{liu} uses matrix factorization-based clustering to assign facet distributions, independently sampling facets for target and context nodes. Splitter \cite{epasto} generates multi-facet node embeddings aligned with the original representation. Asp2vec \cite{park2020unsupervised} introduces a differentiable facet selection module and regularization to explore interactions across facets. For multi-relational graphs, r-GAT \cite{r-GAT} aggregates relation-specific neighborhood information and applies query-aware attention to select relevant facets. However, existing methods are limited to single-node facets in homogeneous graphs. We extend these approaches to heterogeneous graphs by transforming meta-path schemas into diverse facets, linking terminal node facets for richer interaction understanding.

%% file: 7.Conclusion.tex
\section{Conclusion}
We introduced MF2Vec, a new heterogeneous graph neural network that learns node representations using multi-facet paths, independent of node types. MF2Vec consistently outperforms existing methods in node classification, link prediction, and clustering across six benchmark datasets. It demonstrates high stability against meta-path variations and offers improved computational efficiency. Our ablation results confirm the effectiveness of learnable facet selection. In future work, we plan to automate facet and path selection to further improve adaptability and performance. While MF2Vec shows strong generalization and scalability, its current path selection is based on random walks, which may miss some informative paths. Addressing this limitation is an important direction for future research. MF2Vec offers a simple and scalable solution for heterogeneous graph learning.

\begin{acks}
This research was supported by National Research
Foundation of Korea (NRF-2022M3J6A1063021).
\end{acks}

\section*{GenAI Usage Disclosure}
The authors used Generative AI tools (ChatGPT, OpenAI) only for minor grammar checking and assistance in generating table formatting. 
No part of the scientific content, analysis, or results was produced by GenAI.

%% file: main.bbl

\begin{thebibliography}{28}


\ifx \showCODEN    \undefined \def \showCODEN     #1{\unskip}     \fi
\ifx \showDOI      \undefined \def \showDOI       #1{#1}\fi
\ifx \showISBNx    \undefined \def \showISBNx     #1{\unskip}     \fi
\ifx \showISBNxiii \undefined \def \showISBNxiii  #1{\unskip}     \fi
\ifx \showISSN     \undefined \def \showISSN      #1{\unskip}     \fi
\ifx \showLCCN     \undefined \def \showLCCN      #1{\unskip}     \fi
\ifx \shownote     \undefined \def \shownote      #1{#1}          \fi
\ifx \showarticletitle \undefined \def \showarticletitle #1{#1}   \fi
\ifx \showURL      \undefined \def \showURL       {\relax}        \fi
\providecommand\bibfield[2]{#2}
\providecommand\bibinfo[2]{#2}
\providecommand\natexlab[1]{#1}
\providecommand\showeprint[2][]{arXiv:#2}

\bibitem[Bengio et~al\mbox{.}(2013)]%
        {bengio2013estimating}
\bibfield{author}{\bibinfo{person}{Yoshua Bengio}, \bibinfo{person}{Nicholas Léonard}, {and} \bibinfo{person}{Aaron Courville}.} \bibinfo{year}{2013}\natexlab{}.
\newblock \showarticletitle{Estimating or propagating gradients through stochastic neurons for conditional computation}. In \bibinfo{booktitle}{\emph{arXiv preprint arXiv:1308.3432}}.
\newblock


\bibitem[Busbridge et~al\mbox{.}(2019)]%
        {RGAT}
\bibfield{author}{\bibinfo{person}{Dan Busbridge}, \bibinfo{person}{Dane Sherburn}, \bibinfo{person}{Pietro Cavallo}, {and} \bibinfo{person}{Nils~Y. Hammerla}.} \bibinfo{year}{2019}\natexlab{}.
\newblock \showarticletitle{Relational Graph Attention Networks}.
\newblock \bibinfo{journal}{\emph{arXiv preprint arXiv:1904.05811}} (\bibinfo{year}{2019}).
\newblock


\bibitem[Chen et~al\mbox{.}(2019)]%
        {chen2019rgcn}
\bibfield{author}{\bibinfo{person}{Junjie Chen}, \bibinfo{person}{Hongxu Hou}, \bibinfo{person}{Jing Gao}, \bibinfo{person}{Yatu Ji}, {and} \bibinfo{person}{Tiangang Bai}.} \bibinfo{year}{2019}\natexlab{}.
\newblock \showarticletitle{RGCN: recurrent graph convolutional networks for target-dependent sentiment analysis}. In \bibinfo{booktitle}{\emph{International Conference on Knowledge Science, Engineering and Management}}. Springer, \bibinfo{pages}{667--675}.
\newblock


\bibitem[Chen et~al\mbox{.}(2021)]%
        {r-GAT}
\bibfield{author}{\bibinfo{person}{M. Chen}, \bibinfo{person}{Y. Zhang}, \bibinfo{person}{X. Kou}, \bibinfo{person}{Y. Li}, {and} \bibinfo{person}{Y. Zhang}.} \bibinfo{year}{2021}\natexlab{}.
\newblock \showarticletitle{R-GAT: Relational Graph Attention Network for Multi-Relational Graphs}.
\newblock \bibinfo{journal}{\emph{arXiv preprint arXiv:2109.05922}} (\bibinfo{year}{2021}).
\newblock


\bibitem[Defferrard et~al\mbox{.}(2016)]%
        {defferrard}
\bibfield{author}{\bibinfo{person}{Micha{\"e}l Defferrard}, \bibinfo{person}{Xavier Bresson}, {and} \bibinfo{person}{Pierre Vandergheynst}.} \bibinfo{year}{2016}\natexlab{}.
\newblock \showarticletitle{Convolutional neural networks on graphs with fast localized spectral filtering}.
\newblock \bibinfo{journal}{\emph{Advances in Neural Information Processing Systems}}  \bibinfo{volume}{29} (\bibinfo{year}{2016}).
\newblock


\bibitem[Dong et~al\mbox{.}(2017)]%
        {metapath2vec}
\bibfield{author}{\bibinfo{person}{Yuxiao Dong}, \bibinfo{person}{Nitesh~V. Chawla}, {and} \bibinfo{person}{Ananthram Swami}.} \bibinfo{year}{2017}\natexlab{}.
\newblock \showarticletitle{metapath2vec: Scalable representation learning for heterogeneous networks}. In \bibinfo{booktitle}{\emph{Proceedings of the 23rd ACM SIGKDD International Conference on Knowledge Discovery and Data Mining}}. \bibinfo{pages}{135--144}.
\newblock


\bibitem[Epasto and Perozzi(2019)]%
        {epasto}
\bibfield{author}{\bibinfo{person}{Alessandro Epasto} {and} \bibinfo{person}{Bryan Perozzi}.} \bibinfo{year}{2019}\natexlab{}.
\newblock \showarticletitle{Is a single embedding enough? learning node representations that capture multiple social contexts}. In \bibinfo{booktitle}{\emph{The World Wide Web Conference}}. \bibinfo{pages}{394--404}.
\newblock


\bibitem[Fu et~al\mbox{.}(2017)]%
        {hin2vec}
\bibfield{author}{\bibinfo{person}{Tao-yang Fu}, \bibinfo{person}{Wang-Chien Lee}, {and} \bibinfo{person}{Zhen Lei}.} \bibinfo{year}{2017}\natexlab{}.
\newblock \showarticletitle{Hin2vec: Explore meta-paths in heterogeneous information networks for representation learning}. In \bibinfo{booktitle}{\emph{Proceedings of the 2017 ACM on Conference on Information and Knowledge Management}}. \bibinfo{pages}{1797--1806}.
\newblock


\bibitem[Fu and King(2024)]%
        {mecch}
\bibfield{author}{\bibinfo{person}{Xinyu Fu} {and} \bibinfo{person}{Irwin King}.} \bibinfo{year}{2024}\natexlab{}.
\newblock \showarticletitle{MECCH: metapath context convolution-based heterogeneous graph neural networks}.
\newblock \bibinfo{journal}{\emph{Neural Networks}}  \bibinfo{volume}{170} (\bibinfo{year}{2024}), \bibinfo{pages}{266--275}.
\newblock


\bibitem[Fu et~al\mbox{.}(2020)]%
        {magnn}
\bibfield{author}{\bibinfo{person}{Xinyu Fu}, \bibinfo{person}{Jiani Zhang}, \bibinfo{person}{Ziqiao Meng}, {and} \bibinfo{person}{Irwin King}.} \bibinfo{year}{2020}\natexlab{}.
\newblock \showarticletitle{MAGNN: Metapath aggregated graph neural network for heterogeneous graph embedding}. In \bibinfo{booktitle}{\emph{Proceedings of The Web Conference 2020}}. \bibinfo{pages}{2331--2341}.
\newblock


\bibitem[Guan et~al\mbox{.}(2023)]%
        {hmsg}
\bibfield{author}{\bibinfo{person}{Mengya Guan}, \bibinfo{person}{Xinjun Cai}, \bibinfo{person}{Jiaxing Shang}, \bibinfo{person}{Fei Hao}, \bibinfo{person}{Dajiang Liu}, \bibinfo{person}{Xianlong Jiao}, {and} \bibinfo{person}{Wancheng Ni}.} \bibinfo{year}{2023}\natexlab{}.
\newblock \showarticletitle{HMSG: Heterogeneous graph neural network based on Metapath SubGraph learning}.
\newblock \bibinfo{journal}{\emph{Knowledge-Based Systems}}  \bibinfo{volume}{279} (\bibinfo{year}{2023}), \bibinfo{pages}{110930}.
\newblock


\bibitem[Hamilton et~al\mbox{.}(2017)]%
        {hamilton}
\bibfield{author}{\bibinfo{person}{Will Hamilton}, \bibinfo{person}{Zhitao Ying}, {and} \bibinfo{person}{Jure Leskovec}.} \bibinfo{year}{2017}\natexlab{}.
\newblock \showarticletitle{Inductive representation learning on large graphs}.
\newblock \bibinfo{journal}{\emph{Advances in Neural Information Processing Systems}}  \bibinfo{volume}{30} (\bibinfo{year}{2017}).
\newblock


\bibitem[Hong et~al\mbox{.}(2020)]%
        {HetSANN}
\bibfield{author}{\bibinfo{person}{Huiting Hong}, \bibinfo{person}{Hantao Guo}, \bibinfo{person}{Yucheng Lin}, \bibinfo{person}{Xiaoqing Yang}, \bibinfo{person}{Zang Li}, {and} \bibinfo{person}{Jieping Ye}.} \bibinfo{year}{2020}\natexlab{}.
\newblock \showarticletitle{An attention-based graph neural network for heterogeneous structural learning}. In \bibinfo{booktitle}{\emph{Proceedings of the AAAI Conference on Artificial Intelligence}}, Vol.~\bibinfo{volume}{34}. \bibinfo{pages}{4132--4139}.
\newblock


\bibitem[Hu et~al\mbox{.}(2020)]%
        {hu2020heterogeneous}
\bibfield{author}{\bibinfo{person}{Ziniu Hu}, \bibinfo{person}{Yuxiao Dong}, \bibinfo{person}{Kuansan Wang}, {and} \bibinfo{person}{Yizhou Sun}.} \bibinfo{year}{2020}\natexlab{}.
\newblock \showarticletitle{Heterogeneous graph transformer}. In \bibinfo{booktitle}{\emph{Proceedings of the Web Conference 2020}}. \bibinfo{pages}{2704--2710}.
\newblock


\bibitem[Jang et~al\mbox{.}(2016)]%
        {jang2016}
\bibfield{author}{\bibinfo{person}{Eric Jang}, \bibinfo{person}{Shixiang Gu}, {and} \bibinfo{person}{Ben Poole}.} \bibinfo{year}{2016}\natexlab{}.
\newblock \showarticletitle{Categorical reparameterization with gumbel-softmax}.
\newblock \bibinfo{journal}{\emph{arXiv preprint arXiv:1611.01144}} (\bibinfo{year}{2016}).
\newblock


\bibitem[Jin et~al\mbox{.}(2022)]%
        {graphhinge}
\bibfield{author}{\bibinfo{person}{Jiarui Jin}, \bibinfo{person}{Kounianhua Du}, \bibinfo{person}{Weinan Zhang}, \bibinfo{person}{Jiarui Qin}, \bibinfo{person}{Yuchen Fang}, \bibinfo{person}{Yong Yu}, \bibinfo{person}{Zheng Zhang}, {and} \bibinfo{person}{Alexander~J. Smola}.} \bibinfo{year}{2022}\natexlab{}.
\newblock \showarticletitle{GraphHINGE: Learning Interaction Models of Structured Neighborhood on Heterogeneous Information Network}.
\newblock \bibinfo{journal}{\emph{ACM Transactions on Information Systems (TOIS)}} \bibinfo{volume}{40}, \bibinfo{number}{3} (\bibinfo{year}{2022}), \bibinfo{pages}{1--35}.
\newblock


\bibitem[Kipf and Welling(2016)]%
        {kipf}
\bibfield{author}{\bibinfo{person}{Thomas~N. Kipf} {and} \bibinfo{person}{Max Welling}.} \bibinfo{year}{2016}\natexlab{}.
\newblock \showarticletitle{Semi-supervised classification with graph convolutional networks}.
\newblock \bibinfo{journal}{\emph{arXiv preprint arXiv:1609.02907}} (\bibinfo{year}{2016}).
\newblock


\bibitem[Li et~al\mbox{.}(2018)]%
        {LiQ}
\bibfield{author}{\bibinfo{person}{Q. Li}, \bibinfo{person}{Z. Han}, {and} \bibinfo{person}{X.~M. Wu}.} \bibinfo{year}{2018}\natexlab{}.
\newblock \showarticletitle{Deeper Insights into Graph Convolutional Networks for Semi-Supervised Learning}. In \bibinfo{booktitle}{\emph{Proceedings of the AAAI Conference on Artificial Intelligence}}, Vol.~\bibinfo{volume}{32}.
\newblock


\bibitem[Liu et~al\mbox{.}(2019)]%
        {liu}
\bibfield{author}{\bibinfo{person}{Ninghao Liu}, \bibinfo{person}{Qiaoyu Tan}, \bibinfo{person}{Yuening Li}, \bibinfo{person}{Hongxia Yang}, \bibinfo{person}{Jingren Zhou}, {and} \bibinfo{person}{Xia Hu}.} \bibinfo{year}{2019}\natexlab{}.
\newblock \showarticletitle{Is a single vector enough? exploring node polysemy for network embedding}. In \bibinfo{booktitle}{\emph{Proceedings of the 25th ACM SIGKDD International Conference on Knowledge Discovery \& Data Mining}}. \bibinfo{pages}{932--940}.
\newblock


\bibitem[Lv et~al\mbox{.}(2021)]%
        {SimpleHGN}
\bibfield{author}{\bibinfo{person}{Qingsong Lv}, \bibinfo{person}{Ming Ding}, \bibinfo{person}{Qiang Liu}, \bibinfo{person}{Yuxiang Chen}, \bibinfo{person}{Wenzheng Feng}, \bibinfo{person}{Siming He}, \bibinfo{person}{Chang Zhou}, \bibinfo{person}{Jianguo Jiang}, \bibinfo{person}{Yuxiao Dong}, {and} \bibinfo{person}{Jie Tang}.} \bibinfo{year}{2021}\natexlab{}.
\newblock \showarticletitle{Are we really making much progress? revisiting, benchmarking and refining heterogeneous graph neural networks}. In \bibinfo{booktitle}{\emph{Proceedings of the 27th ACM SIGKDD Conference on Knowledge Discovery \& Data Mining}}. \bibinfo{pages}{1150--1160}.
\newblock


\bibitem[Mikolov et~al\mbox{.}(2013)]%
        {skip-gram}
\bibfield{author}{\bibinfo{person}{Tomas Mikolov}, \bibinfo{person}{Kai Chen}, \bibinfo{person}{Greg Corrado}, {and} \bibinfo{person}{Jeffrey Dean}.} \bibinfo{year}{2013}\natexlab{}.
\newblock \showarticletitle{Efficient estimation of word representations in vector space}.
\newblock \bibinfo{journal}{\emph{arXiv preprint arXiv:1301.3781}} (\bibinfo{year}{2013}).
\newblock


\bibitem[Park et~al\mbox{.}(2020)]%
        {park2020unsupervised}
\bibfield{author}{\bibinfo{person}{Chanyoung Park}, \bibinfo{person}{Carl Yang}, \bibinfo{person}{Qi Zhu}, \bibinfo{person}{Donghyun Kim}, \bibinfo{person}{Hwanjo Yu}, {and} \bibinfo{person}{Jiawei Han}.} \bibinfo{year}{2020}\natexlab{}.
\newblock \showarticletitle{Unsupervised differentiable multi-aspect network embedding}. In \bibinfo{booktitle}{\emph{Proceedings of the 26th ACM SIGKDD International Conference on Knowledge Discovery \& Data Mining}}. \bibinfo{pages}{1435--1445}.
\newblock


\bibitem[Sun et~al\mbox{.}(2022)]%
        {mhnf}
\bibfield{author}{\bibinfo{person}{Yundong Sun}, \bibinfo{person}{Dongjie Zhu}, \bibinfo{person}{Haiwen Du}, {and} \bibinfo{person}{Zhaoshuo Tian}.} \bibinfo{year}{2022}\natexlab{}.
\newblock \bibinfo{title}{MHNF: Multi-hop Heterogeneous Neighborhood information Fusion graph representation learning}.
\newblock
\newblock
\showeprint[arxiv]{2106.09289}~[cs.LG]


\bibitem[Van~der Maaten and Hinton(2008)]%
        {t-SNE}
\bibfield{author}{\bibinfo{person}{Laurens Van~der Maaten} {and} \bibinfo{person}{Geoffrey Hinton}.} \bibinfo{year}{2008}\natexlab{}.
\newblock \showarticletitle{Visualizing data using t-SNE}.
\newblock \bibinfo{journal}{\emph{Journal of Machine Learning Research}} \bibinfo{volume}{9}, \bibinfo{number}{11} (\bibinfo{year}{2008}).
\newblock


\bibitem[Velickovic et~al\mbox{.}(2017)]%
        {velickovic}
\bibfield{author}{\bibinfo{person}{Petar Velickovic}, \bibinfo{person}{Guillem Cucurull}, \bibinfo{person}{Arantxa Casanova}, \bibinfo{person}{Adriana Romero}, \bibinfo{person}{Pietro Lio}, {and} \bibinfo{person}{Yoshua Bengio}.} \bibinfo{year}{2017}\natexlab{}.
\newblock \showarticletitle{Graph attention networks}.
\newblock \bibinfo{journal}{\emph{stat}} \bibinfo{volume}{1050}, \bibinfo{number}{20} (\bibinfo{year}{2017}), \bibinfo{pages}{10--48550}.
\newblock


\bibitem[Wang et~al\mbox{.}(2019)]%
        {han}
\bibfield{author}{\bibinfo{person}{Xiao Wang}, \bibinfo{person}{Houye Ji}, \bibinfo{person}{Chuan Shi}, \bibinfo{person}{Bai Wang}, \bibinfo{person}{Yanfang Ye}, \bibinfo{person}{Peng Cui}, {and} \bibinfo{person}{Philip~S. Yu}.} \bibinfo{year}{2019}\natexlab{}.
\newblock \showarticletitle{Heterogeneous graph attention network}. In \bibinfo{booktitle}{\emph{The World Wide Web Conference}}. \bibinfo{pages}{2022--2032}.
\newblock


\bibitem[Wang et~al\mbox{.}(2021)]%
        {wang2021self}
\bibfield{author}{\bibinfo{person}{Xiao Wang}, \bibinfo{person}{Nian Liu}, \bibinfo{person}{Hui Han}, {and} \bibinfo{person}{Chuan Shi}.} \bibinfo{year}{2021}\natexlab{}.
\newblock \showarticletitle{Self-supervised heterogeneous graph neural network with co-contrastive learning}. In \bibinfo{booktitle}{\emph{Proceedings of the 27th ACM SIGKDD Conference on Knowledge Discovery \& Data Mining}}. \bibinfo{pages}{1726--1736}.
\newblock


\bibitem[Yun et~al\mbox{.}(2019)]%
        {GTN}
\bibfield{author}{\bibinfo{person}{S. Yun}, \bibinfo{person}{M. Jeong}, \bibinfo{person}{R. Kim}, \bibinfo{person}{J. Kang}, {and} \bibinfo{person}{H.~J. Kim}.} \bibinfo{year}{2019}\natexlab{}.
\newblock \showarticletitle{Graph Transformer Networks}. In \bibinfo{booktitle}{\emph{Advances in Neural Information Processing Systems}}, Vol.~\bibinfo{volume}{39}. \bibinfo{pages}{1--12}.
\newblock


\end{thebibliography}
